# A Model of Inductive Bias Learning

**Jonathan Baxter**                                        Jonathan.Baxter@anu.edu.au
*Research School of Information Sciences and Engineering*
*Australian National University, Canberra 0200, Australia*

## Abstract

A major problem in machine learning is that of inductive bias: how to choose a learner's hypothesis space so that it is large enough to contain a solution to the problem being learnt, yet small enough to ensure reliable generalization from reasonably-sized training sets. Typically such bias is supplied by hand through the skill and insights of experts. In this paper a model for *automatically learning* bias is investigated. The central assumption of the model is that the learner is embedded within an *environment* of related learning tasks. Within such an environment the learner can sample from multiple tasks, and hence it can search for a hypothesis space that contains good solutions to many of the problems in the environment. Under certain restrictions on the set of all hypothesis spaces available to the learner, we show that a hypothesis space that performs well on a sufficiently large number of training tasks will also perform well when learning novel tasks in the same environment. Explicit bounds are also derived demonstrating that learning multiple tasks within an environment of related tasks can potentially give much better generalization than learning a single task.

## 1. Introduction

Often the hardest problem in any machine learning task is the initial choice of hypothesis space; it has to be large enough to contain a solution to the problem at hand, yet small enough to ensure good generalization from a small number of examples (Mitchell, 1991). Once a suitable bias has been found, the actual learning task is often straightforward. Existing methods of bias generally require the input of a human expert in the form of heuristics and domain knowledge (for example, through the selection of an appropriate set of features). Despite their successes, such methods are clearly limited by the accuracy and reliability of the expert's knowledge and also by the extent to which that knowledge can be transferred to the learner. Thus it is natural to search for methods for *automatically learning* the bias.

In this paper we introduce and analyze a formal model of *bias learning* that builds upon the PAC model of machine learning and its variants (Vapnik, 1982; Valiant, 1984; Blumer, Ehrenfeucht, Haussler, & Warmuth, 1989; Haussler, 1992). These models typically take the following general form: the learner is supplied with a hypothesis space $\mathcal{H}$ and training data $z = \{(x_1, y_1), \ldots, (x_m, y_m)\}$ drawn independently according to some underlying distribution $P$ on $X \times Y$. Based on the information contained in $z$, the learner's goal is to select a hypothesis $h \colon X \to Y$ from $\mathcal{H}$ minimizing some measure $\mathrm{er}_P(h)$ of expected loss with respect to $P$ (for example, in the case of squared loss $\mathrm{er}_P(h) := \mathbb{E}_{(x,y) \sim P} \left( h(x) - y \right)^2$). In such models the learner's bias is represented by the choice of $\mathcal{H}$; if $\mathcal{H}$ does not contain a good solution to the problem, then, regardless of how much data the learner receives, it cannot learn.

Of course, the best way to bias the learner is to supply it with an $\mathcal{H}$ containing just a single optimal hypothesis. But finding such a hypothesis is precisely the original learning problem, so in the PAC model there is no distinction between bias learning and ordinary learning. Or put differently, the PAC model does not model the process of inductive bias, it simply takes the hypothesis space $\mathcal{H}$ as given and proceeds from there. To overcome this problem, in this paper we assume that instead of being faced with just a single learning task, the learner is embedded within an *environment* of related learning tasks. The learner is supplied with a *family* of hypothesis spaces $\mathbb{H} = \{\mathcal{H}\}$, and its goal is to find a bias (i.e. hypothesis space $\mathcal{H} \in \mathbb{H}$) that is appropriate for the entire environment. A simple example is the problem of handwritten character recognition. A preprocessing stage that identifies and removes any (small) rotations, dilations and translations of an image of a character will be advantageous for recognizing all characters. If the set of all individual character recognition problems is viewed as an environment of learning problems (that is, the set of all problems of the form "distinguish 'A' from all other characters", "distinguish 'B' from all other characters", and so on), this preprocessor represents a bias that is appropriate for all problems in the environment. It is likely that there are many other currently unknown biases that are also appropriate for this environment. We would like to be able to learn these automatically.

There are many other examples of learning problems that can be viewed as belonging to environments of related problems. For example, each individual face recognition problem belongs to an (essentially infinite) set of related learning problems (all the other individual face recognition problems); the set of all individual spoken word recognition problems forms another large environment, as does the set of all fingerprint recognition problems, printed Chinese and Japanese character recognition problems, stock price prediction problems and so on. Even medical diagnostic and prognostic problems, where a multitude of diseases are predicted from the same pathology tests, constitute an environment of related learning problems.

In many cases these "environments" are not normally modeled as such; instead they are treated as single, multiple category learning problems. For example, recognizing a group of faces would normally be viewed as a single learning problem with multiple class labels (one for each face in the group), not as multiple individual learning problems. However, if a reliable classifier for each individual face in the group can be constructed then they can easily be combined to produce a classifier for the whole group. Furthermore, by viewing the faces as an environment of related learning problems, the results presented here show that bias can be learnt that will be good for learning novel faces, a claim that cannot be made for the traditional approach.

This point goes to the heart of our model: we are not not concerned with adjusting a learner's bias so it performs better on some *fixed* set of learning problems. Such a process is in fact just ordinary learning but with a richer hypothesis space in which some components labelled "bias" are also able to be varied. Instead, we suppose the learner is faced with a (potentially infinite) stream of tasks, and that by adjusting its bias on some subset of the tasks it improves its learning performance on future, as yet unseen tasks.

Bias that is appropriate for all problems in an environment must be learnt by sampling from many tasks. If only a single task is learnt then the bias extracted is likely to be specific to that task. In the rest of this paper, a general theory of bias learning is developed based upon the idea of learning multiple related tasks. Loosely speaking (formal results are stated in Section 2), there are two main conclusions of the theory presented here:

- Learning multiple related tasks reduces the sampling burden required for good generalization, at least on a number-of-examples-required-per-task basis.

- Bias that is learnt on sufficiently many training tasks is likely to be good for learning novel tasks drawn from the same environment.

The second point shows that a form of *meta-generalization* is possible in bias learning. Ordinarily, we say a learner generalizes well if, after seeing sufficiently many training examples, it produces a hypothesis that with high probability will perform well on future examples of the same task. However, a bias learner generalizes well if, after seeing sufficiently many training *tasks* it produces a *hypothesis space* that with high probability contains good solutions to novel tasks. Another term that has been used for this process is *Learning to Learn* (Thrun & Pratt, 1997).

Our main theorems are stated in an agnostic setting (that is, $\mathbb{H}$ does not necessarily contain a hypothesis space with solutions to all the problems in the environment), but we also give improved bounds in the realizable case. The sample complexity bounds appearing in these results are stated in terms of combinatorial parameters related to the complexity of the set of all hypothesis spaces $\mathbb{H}$ available to the bias learner. For Boolean learning problems (pattern classification) these parameters are the bias learning analogue of the *Vapnik-Chervonenkis dimension* (Vapnik, 1982; Blumer et al., 1989).

As an application of the general theory, the problem of learning an appropriate set of neural-network features for an environment of related tasks is formulated as a bias learning problem. In the case of continuous neural-network features we are able to prove upper bounds on the number of training tasks and number of examples of each training task required to ensure a set of features that works well for the training tasks will, with high probability, work well on novel tasks drawn from the same environment. The upper bound on the number of tasks scales as $O(b)$ where $b$ is a measure of the complexity of the possible feature sets available to the learner, while the upper bound on the number of examples of each task scales as $O(a + b/n)$ where $O(a)$ is the number of examples required to learn a task if the "true" set of features (that is, the correct bias) is already known, and $n$ is the number of tasks. Thus, in this case we see that as the number of related tasks learnt increases, the number of examples required of each task for good generalization decays to the minimum possible. For Boolean neural-network feature maps we are able to show a matching lower bound on the number of examples required per task of the same form.

## 1.1 Related Work

There is a large body of previous algorithmic and experimental work in the machine learning and statistics literature addressing the problems of inductive bias learning and improving generalization through multiple task learning. Some of these approaches can be seen as special cases of, or at least closely aligned with, the model described here, while others are more orthogonal. Without being completely exhaustive, in this section we present an overview of the main contributions. See Thrun and Pratt (1997, chapter 1) for a more comprehensive treatment.

- **Hierarchical Bayes.** The earliest approaches to bias learning come from Hierarchical Bayesian methods in statistics (Berger, 1985; Good, 1980; Gelman, Carlin, Stern, & Rubim, 1995). In contrast to the Bayesian methodology, the present paper takes an essentially empirical process approach to modeling the problem of bias learning. However, a model using a mixture of hierarchical Bayesian and information-theoretic ideas was presented in Baxter (1997a), with similar conclusions to those found here. An empirical study showing the utility of the hierarchical Bayes approach in a domain containing a large number of related tasks was given in Heskes (1998).

- **Early machine learning work.** In Rendell, Seshu, and Tcheng (1987) "VBMS" or *Variable Bias Management System* was introduced as a mechanism for selecting amongst different learning algorithms when tackling a new learning problem. "STABB" or *Shift To a Better Bias* (Utgoff, 1986) was another early scheme for adjusting bias, but unlike VBMS, STABB was not primarily focussed on searching for bias applicable to large problem domains. Our use of an "environment of related tasks" in this paper may also be interpreted as an "environment of analogous tasks" in the sense that conclusions about one task can be arrived at by analogy with (sufficiently many of) the other tasks. For an early discussion of analogy in this context, see Russell (1989, S4.3), in particular the observation that for analogous problems the sampling burden *per task* can be reduced.

- **Metric-based approaches.** The metric used in nearest-neighbour classification, and in vector quantization to determine the nearest code-book vector, represents a form of inductive bias. Using the model of the present paper, and under some extra assumptions on the tasks in the environment (specifically, that their marginal input-space distributions are identical and they only differ in the conditional probabilities they assign to class labels), it can be shown that there is an *optimal* metric or distance measure to use for vector quantization and one-nearest-neighbour classification (Baxter, 1995a, 1997b; Baxter & Bartlett, 1998). This metric can be learnt by sampling from a subset of tasks from the environment, and then used as a distance measure when learning novel tasks drawn from the same environment. Bounds on the number of tasks and examples of each task required to ensure good performance on novel tasks were given in Baxter and Bartlett (1998), along with an experiment in which a metric was successfully trained on examples of a subset of 400 Japanese characters and then used as a fixed distance measure when learning 2600 as yet unseen characters.

  A similar approach is described in Thrun and Mitchell (1995), Thrun (1996), in which a neural network's output was trained to match labels on a novel task, while simultaneously being forced to match its gradient to *derivative* information generated from a distance metric trained on previous, related tasks. Performance on the novel tasks improved substantially with the use of the derivative information.

  Note that there are many other adaptive metric techniques used in machine learning, but these all focus exclusively on adjusting the metric for a fixed set of problems rather than learning a metric suitable for learning novel, related tasks (bias learning).

- **Feature learning or learning internal representations.** As with adaptive metric techniques, there are many approaches to feature learning that focus on adapting features for a fixed task rather than learning features to be used in novel tasks. One of the few cases where features have been learnt on a subset of tasks with the explicit aim of using them on novel tasks was Intrator and Edelman (1996) in which a low-dimensional representation was learnt for a set of multiple related image-recognition tasks and then used to successfully learn novel tasks of the same kind. The experiments reported in Baxter (1995a, chapter 4) and Baxter (1995b), Baxter and Bartlett (1998) are also of this nature.

- **Bias learning in Inductive Logic Programming (ILP).** Predicate invention refers to the process in ILP whereby new predicates thought to be useful for the classification task at hand are added to the learner's domain knowledge. By using the new predicates as background domain knowledge when learning novel tasks, predicate invention may be viewed as a form of inductive bias learning. Preliminary results with this approach on a chess domain are reported in Khan, Muggleton, and Parson (1998).

- **Improving performance on a fixed reference task.** "Multi-task learning" (Caruana, 1997) trains extra neural network outputs to match related tasks in order to improve generalization performance on a fixed reference task. Although this approach does not explicitly identify the extra bias generated by the related tasks in a way that can be used to learn novel tasks, it is an example of exploiting the bias provided by a set of related tasks to improve generalization performance. Other similar approaches include Suddarth and Kergosien (1990), Suddarth and Holden (1991), Abu-Mostafa (1993).

- **Bias as computational complexity.** In this paper we consider inductive bias from a sample-complexity perspective: how does the learnt bias decrease the number of examples required of novel tasks for good generalization? A natural alternative line of enquiry is how the running-time or computational complexity of a learning algorithm may be improved by training on related tasks. Some early algorithms for neural networks in this vein are contained in Sharkey and Sharkey (1993), Pratt (1992).

- **Reinforcement Learning.** Many control tasks can appropriately be viewed as elements of sets of related tasks, such as learning to navigate to different goal states, or learning a set of complex motor control tasks. A number of papers in the reinforcement learning literature have proposed algorithms for both sharing the information in related tasks to improve average generalization performance across those tasks Singh (1992), Ring (1995), or learning bias from a set of tasks to improve performance on future tasks Sutton (1992), Thrun and Schwartz (1995).

## 1.2 Overview of the Paper

In Section 2 the bias learning model is formally defined, and the main sample complexity results are given showing the utility of learning multiple related tasks and the feasibility of bias learning. These results show that the sample complexity is controlled by the size of certain covering numbers associated with the set of all hypothesis spaces available to the bias learner, in much the same way as the sample complexity in learning Boolean functions is controlled by the *Vapnik-Chervonenkis* dimension (Vapnik, 1982; Blumer et al., 1989). The results of Section 2 are upper bounds on the sample complexity required for good generalization when learning multiple tasks and learning inductive bias.

The general results of Section 2 are specialized to the case of feature learning with neural networks in Section 3, where an algorithm for training features by gradient descent is also presented. For this special case we are able to show matching lower bounds for the sample complexity of multiple task learning. In Section 4 we present some concluding remarks and directions for future research. Many of the proofs are quite lengthy and have been moved to the appendices so as not to interrupt the flow of the main text.

The following tables contain a glossary of the mathematical symbols used in the paper.

| Symbol | Description | First Referenced |
|---|---|---|
| $X$ | Input Space | 155 |
| $Y$ | Output Space | 155 |
| $P$ | Distribution on $X \times Y$ (learning task) | 155 |
| $l$ | Loss function | 155 |
| $\mathcal{H}$ | Hypothesis Space | 155 |
| $h$ | Hypothesis | 155 |
| $\mathrm{er}_P(h)$ | Error of hypothesis $h$ on distribution $P$ | 156 |
| $z$ | Training set | 156 |
| $\mathcal{A}$ | Learning Algorithm | 156 |
| $\hat{\mathrm{er}}_z(h)$ | Empirical error of $h$ on training set $z$ | 156 |
| $\mathcal{P}$ | Set of all learning tasks $P$ | 157 |
| $Q$ | Distribution over learning tasks | 157 |
| $\mathbb{H}$ | Family of hypothesis spaces | 157 |
| $\mathrm{er}_Q(\mathcal{H})$ | Loss of hypothesis space $\mathcal{H}$ on environment $Q$ | 158 |
| $\mathbf{z}$ | $(n, m)$-sample | 158 |
| $\hat{\mathrm{er}}_\mathbf{z}(\mathcal{H})$ | Empirical loss of $\mathcal{H}$ on $\mathbf{z}$ | 158 |
| $\mathcal{A}$ | Bias learning algorithm | 159 |
| $h_l$ | Function induced by $h$ and $l$ | 159 |
| $\mathcal{H}_l$ | Set of $h_l$ | 159 |
| $(h_1, \ldots, h_n)_l$ | Average of $h_{1,l}, \ldots, h_{n,l}$ | 159 |
| $\mathbf{h}_l$ | Same as $(h_1, \ldots, h_n)_l$ | 159 |
| $\mathcal{H}_l^n$ | Set of $(h_1, \ldots, h_n)_l$ | 159 |
| $\mathbb{H}_l^n$ | Set of $\mathcal{H}_l^n$ | 159 |
| $\mathcal{H}^*$ | Function on probability distributions | 160 |
| $\mathbb{H}^*$ | Set of $\mathcal{H}^*$ | 160 |
| $d_\mathbf{P}$ | Pseudo-metric on $\mathcal{H}_l^n$ | 160 |
| $d_Q$ | Pseudo-metric on $\mathbb{H}^*$ | 160 |
| $\mathcal{N}(\varepsilon, \mathbb{H}^*, d_Q)$ | Covering number of $\mathbb{H}^*$ | 160 |
| $\mathcal{C}(\varepsilon, \mathbb{H}^*)$ | Capacity of $\mathbb{H}^*$ | 160 |
| $\mathcal{N}(\varepsilon, \mathbb{H}_l^n, d_\mathbf{P})$ | Covering number of $\mathbb{H}_l^n$ | 160 |
| $C(\varepsilon, \mathbb{H}_l^n)$ | Capacity of $\mathbb{H}_l^n$ | 160 |
| $\mathbf{h}$ | Sequence of $n$ hypotheses $(h_1, \ldots, h_n)$ | 163 |
| $\mathbf{P}$ | Sequence of $n$ distributions $(P_1, \ldots, P_n)$ | 163 |
| $\mathrm{er}_\mathbf{P}(\mathbf{h})$ | Average loss of $\mathbf{h}$ on $\mathbf{P}$ | 164 |
| $\hat{\mathrm{er}}_\mathbf{z}(\mathbf{h})$ | Average loss of $\mathbf{h}$ on $\mathbf{z}$ | 164 |
| $\mathcal{F}$ | Set of feature maps | 166 |
| $\mathcal{G}$ | Output class composed with feature maps $f$ | 166 |
| $\mathcal{G} \circ f$ | Hypothesis space associated with $f$ | 166 |
| $\mathcal{G}_l$ | Loss function class associated with $\mathcal{G}$ | 166 |
| $\mathcal{N}(\varepsilon, \mathcal{G}_l, d_P)$ | Covering number of $\mathcal{G}_l$ | 166 |
| $\mathcal{C}(\varepsilon, \mathcal{G}_l)$ | Capacity of $\mathcal{G}_l$ | 166 |
| $d_{[P, \mathcal{G}_l]}(f, f')$ | Pseudo-metric on feature maps $f, f'$ | 166 |
| $\mathcal{N}(\varepsilon, \mathcal{F}, d_{[P, \mathcal{G}_l]})$ | Covering number of $\mathcal{F}$ | 166 |

| Symbol | Description | First Referenced |
|---|---|---|
| $\mathcal{N}(\varepsilon, \mathcal{F}, d_{[P, \mathcal{G}_l]})$ | Covering number of $\mathcal{F}$ | 166 |
| $\mathcal{C}_{\mathcal{G}_l}(\varepsilon, \mathcal{F})$ | Capacity of $\mathcal{F}$ | 166 |
| $\mathcal{H}_w$ | Neural network hypothesis space | 167 |
| $\mathcal{H}_{|x}$ | $\mathcal{H}$ restricted to vector $x$ | 172 |
| $\Pi_{\mathcal{H}}(m)$ | Growth function of $\mathcal{H}$ | 172 |
| $\text{VCdim}(\mathcal{H})$ | Vapnik-Chervonenkis dimension of $\mathcal{H}$ | 172 |
| $\mathcal{H}_{|\mathbf{x}}$ | $\mathcal{H}$ restricted to matrix $\mathbf{x}$ | 173 |
| $\mathbb{H}_{|\mathbf{x}}$ | $\mathbb{H}$ restricted to matrix $\mathbf{x}$ | 173 |
| $\Pi_{\mathbb{H}}(n, m)$ | Growth function of $\mathbb{H}$ | 173 |
| $d_{\mathbb{H}}(n)$ | Dimension function of $\mathbb{H}$ | 173 |
| $\overline{d}(\mathbb{H})$ | Upper dimension function of $\mathbb{H}$ | 173 |
| $\underline{d}(\mathbb{H})$ | Lower dimension function of $\mathbb{H}$ | 173 |
| $\text{opt}_{\mathbf{P}}(\mathbb{H}^n)$ | Optimal performance of $\mathbb{H}^n$ on $\mathbf{P}$ | 175 |
| $d_\nu$ | Metric on $\mathbb{R}^+$ | 179 |
| $h_1 \oplus \cdots \oplus h_n$ | Average of $h_1, \ldots, h_n$ | 179 |
| $\mathcal{H}_1 \oplus \cdots \oplus \mathcal{H}_n$ | Set of $h_1 \oplus \cdots \oplus h_n$ | 180 |
| $\Gamma_{(2m,n)}$ | Permutations on integer pairs | 182 |
| $\mathbf{z}_\sigma$ | Permuted $\mathbf{z}$ | 182 |
| $d_{\mathbf{z}}(\mathbf{h}, \mathbf{h}')$ | Empirical $l_1$ metric on functions $\mathbf{h}$ | 182 |
| $\hat{\text{er}}_{\mathbf{P}}(\mathcal{H})$ | Optimal average error of $\mathcal{H}$ on $\mathbf{P}$ | 185 |

## 2. The Bias Learning Model

In this section the bias learning model is formally introduced. To motivate the definitions, we first describe the main features of ordinary (single-task) supervised learning models.

### 2.1 Single-Task Learning

Computational learning theory models of supervised learning usually include the following ingredients:

- An *input space* $X$ and an *output space* $Y$,

- a *probability distribution* $P$ on $X \times Y$,

- a *loss function* $l: Y \times Y \to \mathbb{R}$, and

- a *hypothesis space* $\mathcal{H}$ which is a set of *hypotheses* or functions $h: X \to Y$.

As an example, if the problem is to learn to recognize images of Mary's face using a neural network, then $X$ would be the set of all images (typically represented as a subset of $\mathbb{R}^d$ where each component is a pixel intensity), $Y$ would be the set $\{0, 1\}$, and the distribution $P$ would be peaked over images of different faces and the correct class labels. The learner's hypothesis space $\mathcal{H}$ would be a class of neural networks mapping the input space $\mathbb{R}^d$ to $\{0, 1\}$. The loss in this case would be discrete loss:

$$l(y, y') := \begin{cases} 1 & \text{if } y \neq y' \\ 0 & \text{if } y = y' \end{cases} \qquad (1)$$

Using the loss function allows us to present a unified treatment of both pattern recognition ($Y = \{0, 1\}$, $l$ as above), and real-valued function learning (*e.g.* regression) in which $Y = \mathbb{R}$ and usually $l(y, y') = (y - y')^2$.

The goal of the learner is to select a hypothesis $h \in \mathcal{H}$ with minimum *expected loss*:

$$\mathrm{er}_P(h) := \int_{X \times Y} l(h(x), y) \, dP(x, y). \tag{2}$$

Of course, the learner does not know $P$ and so it cannot search through $\mathcal{H}$ for an $h$ minimizing $\mathrm{er}_P(h)$. In practice, the learner samples repeatedly from $X \times Y$ according to the distribution $P$ to generate a *training set*

$$z := \{(x_1, y_1), \ldots, (x_m, y_m)\}. \tag{3}$$

Based on the information contained in $z$ the learner produces a hypothesis $h \in \mathcal{H}$. Hence, in general a learner is simply a map $\mathcal{A}$ from the set of all training samples to the hypothesis space $\mathcal{H}$:

$$\mathcal{A} \colon \bigcup_{m>0} (X \times Y)^m \to \mathcal{H}$$

(stochastic learner's can be treated by assuming a distribution-valued $\mathcal{A}$.)

Many algorithms seek to minimize the *empirical* loss of $h$ on $z$, where this is defined by:

$$\hat{\mathrm{er}}_z(h) := \frac{1}{m} \sum_{i=1}^{m} l(h(x_i), y_i). \tag{4}$$

Of course, there are more intelligent things to do with the data than simply minimizing empirical error—for example one can add regularisation terms to avoid over-fitting.

However the learner chooses its hypothesis $h$, if we have a *uniform* bound (over all $h \in \mathcal{H}$) on the probability of large deviation between $\hat{\mathrm{er}}_z(h)$ and $\mathrm{er}_P(h)$, then we can bound the learner's generalization error $\mathrm{er}_P(h)$ as a function of its empirical loss on the training set $\hat{\mathrm{er}}_z(h)$. Whether such a bound holds depends upon the "richness" of $\mathcal{H}$. The conditions ensuring convergence between $\hat{\mathrm{er}}_z(h)$ and $\mathrm{er}_P(h)$ are by now well understood; for Boolean function learning ($Y = \{0, 1\}$, discrete loss), convergence is controlled by the *VC-dimension*[1] of $\mathcal{H}$:

**Theorem 1.** *Let $P$ be any probability distribution on $X \times \{0, 1\}$ and suppose $z = \{(x_1, y_1), \ldots, (x_m, y_m)\}$ is generated by sampling $m$ times from $X \times \{0, 1\}$ according to $P$. Let $d := \mathrm{VCdim}(\mathcal{H})$. Then with probability at least $1 - \delta$ (over the choice of the training set $z$),* **all** *$h \in \mathcal{H}$ will satisfy*

$$\mathrm{er}_P(h) \leq \hat{\mathrm{er}}_z(h) + \left[ \frac{32}{m} \left( d \log \frac{2em}{d} + \log \frac{4}{\delta} \right) \right]^{1/2} \tag{5}$$

Proofs of this result may be found in Vapnik (1982), Blumer et al. (1989), and will not be reproduced here.

---

1. The VC dimension of a class of Boolean functions $\mathcal{H}$ is the largest integer $d$ such that there exists a subset $S := \{x_1, \ldots, x_d\} \subset X$ such that the restriction of $\mathcal{H}$ to $S$ contains all $2^d$ Boolean functions on $S$.

Theorem 1 only provides conditions under which the deviation between $\mathrm{er}_P(h)$ and $\hat{\mathrm{er}}_{\mathbf{z}}(h)$ is likely to be small, it does not guarantee that the true error $\mathrm{er}_P(h)$ will actually be small. This is governed by the choice of $\mathcal{H}$. If $\mathcal{H}$ contains a solution with small error and the learner minimizes error on the training set, then with high probability $\mathrm{er}_P(h)$ will be small. However, a bad choice of $\mathcal{H}$ will mean there is no hope of achieving small error. Thus, the *bias* of the learner in this model[2] is represented by the choice of hypothesis space $\mathcal{H}$.

## 2.2 The Bias Learning Model

The main extra assumption of the bias learning model introduced here is that the learner is embedded in an *environment* of related tasks, and can sample from the environment to generate multiple training sets belonging to multiple different tasks. In the above model of ordinary (single-task) learning, a learning task is represented by a distribution $P$ on $X \times Y$. So in the bias learning model, an environment of learning problems is represented by a pair $(\mathcal{P}, Q)$ where $\mathcal{P}$ is the set of all probability distributions on $X \times Y$ (i.e., $\mathcal{P}$ is the set of all possible learning problems), and $Q$ is a distribution on $\mathcal{P}$. $Q$ controls which learning problems the learner is likely to see[3]. For example, if the learner is in a face recognition environment, $Q$ will be highly peaked over face-recognition-type problems, whereas if the learner is in a character recognition environment $Q$ will be peaked over character-recognition-type problems (here, as in the introduction, we view these environments as sets of individual classification problems, rather than single, multiple class classification problems).

Recall from the last paragraph of the previous section that the learner's bias is represented by its choice of hypothesis space $\mathcal{H}$. So to enable the learner to learn the bias, we supply it with a *family* or set of hypothesis spaces $\mathbb{H} := \{\mathcal{H}\}$.

Putting all this together, formally a *learning to learn* or *bias learning* problem consists of:

- an *input space* $X$ and an *output space* $Y$ (both of which are separable metric spaces),

- a *loss function* $l \colon Y \times Y \to \mathbb{R}$,

- an *environment* $(\mathcal{P}, Q)$ where $\mathcal{P}$ is the set of all probability distributions on $X \times Y$ and $Q$ is a distribution on $\mathcal{P}$,

- a *hypothesis space family* $\mathbb{H} = \{\mathcal{H}\}$ where each $\mathcal{H} \in \mathbb{H}$ is a set of functions $h \colon X \to Y$.

From now on we will assume the loss function $l$ has range $[0, 1]$, or equivalently, with rescaling, we assume that $l$ is bounded.

---

2. The bias is also governed by how the learner uses the hypothesis space. For example, under some circumstances the learner may choose not to use the full power of $\mathcal{H}$ (a neural network example is early-stopping). For simplicity in this paper we abstract away from such features of the algorithm $\mathcal{A}$ and assume that it uses the entire hypothesis space $\mathcal{H}$.

3. $Q$'s domain is a $\sigma$-algebra of subsets of $\mathcal{P}$. A suitable one for our purposes is the Borel $\sigma$-algebra $\mathcal{B}(\mathcal{P})$ generated by the topology of weak convergence on $\mathcal{P}$. If we assume that $X$ and $Y$ are separable metric spaces, then $\mathcal{P}$ is also a separable metric space in the Prohorov metric (which metrizes the topology of weak convergence) (Parthasarathy, 1967), so there is no problem with the existence of measures on $\mathcal{B}(\mathcal{P})$. See Appendix D for further discussion, particularly the proof of part 5 in Lemma 32.

We define the goal of a bias learner to be to find a hypothesis space $\mathcal{H} \in \mathbb{H}$ minimizing the following loss:

$$\mathrm{er}_Q(\mathcal{H}) := \int_{\mathcal{P}} \inf_{h \in \mathcal{H}} \mathrm{er}_P(h) \, dQ(P) \tag{6}$$
$$= \int_{\mathcal{P}} \inf_{h \in \mathcal{H}} \int_{X \times Y} l(h(x), y) \, dP(x, y) \, dQ(P).$$

The only way $\mathrm{er}_Q(\mathcal{H})$ can be small is if, with high $Q$-probability, $\mathcal{H}$ contains a good solution $h$ to any problem $P$ drawn at random according to $Q$. In this sense $\mathrm{er}_Q(\mathcal{H})$ measures how appropriate the bias embodied by $\mathcal{H}$ is for the environment $(\mathcal{P}, Q)$.

In general the learner will not know $Q$, so it will not be able to find an $\mathcal{H}$ minimizing $\mathrm{er}_Q(\mathcal{H})$ directly. However, the learner can sample from the environment in the following way:

- Sample $n$ times from $\mathcal{P}$ according to $Q$ to yield:
  $P_1, \ldots, P_n$.

- Sample $m$ times from $X \times Y$ according to each $P_i$ to yield:
  $z_i = \{(x_{i1}, y_{i1}) \ldots, (x_{im}, y_{im})\}$.

- The resulting $n$ training sets—henceforth called an $(n, m)$-*sample* if they are generated by the above process—are supplied to the learner. In the sequel, an $(n, m)$-sample will be denoted by $\mathbf{z}$ and written as a matrix:

$$\mathbf{z} := \begin{matrix} (x_{11}, y_{11}) & \cdots & (x_{1m}, y_{1m}) & = z_1 \\ \vdots & \ddots & \vdots & \vdots \\ (x_{n1}, y_{n1}) & \cdots & (x_{nm}, y_{nm}) & = z_n \end{matrix} \tag{7}$$

An $(n, m)$-sample is simply $n$ training sets $z_1, \ldots, z_n$ sampled from $n$ different learning tasks $P_1, \ldots, P_n$, where each task is selected according to the environmental probability distribution $Q$. The size of each training set is kept the same primarily to facilitate the analysis.

Based on the information contained in $\mathbf{z}$, the learner must choose a hypothesis space $\mathcal{H} \in \mathbb{H}$. One way to do this would be for the learner to find an $\mathcal{H}$ minimizing the *empirical loss* on $\mathbf{z}$, where this is defined by:

$$\hat{\mathrm{er}}_{\mathbf{z}}(\mathcal{H}) := \frac{1}{n} \sum_{i=1}^{n} \inf_{h \in \mathcal{H}} \hat{\mathrm{er}}_{z_i}(h) \tag{8}$$

Note that $\hat{\mathrm{er}}_{\mathbf{z}}(\mathcal{H})$ is simply the average of the best possible empirical error achievable on each training set $z_i$, using a function from $\mathcal{H}$. It is a biased estimate of $\mathrm{er}_Q(\mathcal{H})$. An unbiased estimate of $\mathrm{er}_Q(\mathcal{H})$ would require choosing an $\mathcal{H}$ with minimal average error over the $n$ distributions $P_1, \ldots, P_n$, where this is defined by $\frac{1}{n} \sum_{i=1}^{n} \inf_{h \in \mathcal{H}} \mathrm{er}_{P_i}(h)$.

As with ordinary learning, it is likely there are more intelligent things to do with the training data $\mathbf{z}$ than minimizing (8). Denoting the set of all $(n, m)$-samples by $(X \times Y)^{(n,m)}$, a general "bias learner" is a map $\mathcal{A}$ that takes $(n, m)$-samples as input and produces hypothesis spaces $\mathcal{H} \in \mathbb{H}$ as output:

$$\mathcal{A} \colon \bigcup_{\substack{n>0 \\ m>0}} (X \times Y)^{(n,m)} \to \mathbb{H}. \tag{9}$$

(as stated, $\mathcal{A}$ is a deterministic bias learner, however it is trivial to extend our results to stochastic learners).

Note that in this paper we are concerned only with the sample complexity properties of a bias learner $\mathcal{A}$; we do not discuss issues of the computability of $\mathcal{A}$.

Since $\mathcal{A}$ is searching for entire hypothesis spaces $\mathcal{H}$ within a family of such hypothesis spaces $\mathbb{H}$, there is an extra representational question in our model of bias learning that is not present in ordinary learning, and that is how the family $\mathbb{H}$ is represented and searched by $\mathcal{A}$. We defer this discussion until Section 2.5, after the main sample complexity results for this model of bias learning have been introduced. For the specific case of learning a set of features suitable for an environment of related learning problems, see Section 3.

Regardless of how the learner chooses its hypothesis space $\mathcal{H}$, if we have a uniform bound (over all $\mathcal{H} \in \mathbb{H}$) on the probability of large deviation between $\hat{\mathrm{er}}_{\mathbf{z}}(\mathcal{H})$ and $\mathrm{er}_Q(\mathcal{H})$, and we can compute an upper bound on $\hat{\mathrm{er}}_{\mathbf{z}}(\mathcal{H})$, then we can bound the bias learner's "generalization error" $\mathrm{er}_Q(\mathcal{H})$. With this view, the question of generalization within our bias learning model becomes: how many tasks ($n$) and how many examples of each task ($m$) are required to ensure that $\hat{\mathrm{er}}_{\mathbf{z}}(\mathcal{H})$ and $\mathrm{er}_Q(\mathcal{H})$ are close with high probability, uniformly over all $\mathcal{H} \in \mathbb{H}$? Or, informally, how many tasks and how many examples of each task are required to ensure that a hypothesis space with good solutions to all the training tasks will contain good solutions to novel tasks drawn from the same environment?

It turns out that this kind of uniform convergence for bias learning is controlled by the "size" of certain function classes derived from the hypothesis space family $\mathbb{H}$, in much the same way as the VC-dimension of a hypothesis space $\mathcal{H}$ controls uniform convergence in the case of Boolean function learning (Theorem 1). These "size" measures and other auxiliary definitions needed to state the main theorem are introduced in the following subsection.

## 2.3 Covering Numbers

**Definition 1.** *For any hypothesis* $h\colon X \to Y$, *define* $h_l\colon X \times Y \to [0,1]$ *by*

$$h_l(x,y) := l(h(x), y) \tag{10}$$

*For any hypothesis space* $\mathcal{H}$ *in the hypothesis space family* $\mathbb{H}$, *define*

$$\mathcal{H}_l := \{h_l \colon h \in \mathcal{H}\}. \tag{11}$$

*For any sequence of* $n$ *hypotheses* $(h_1, \ldots, h_n)$, *define* $(h_1, \ldots, h_n)_l \colon (X \times Y)^n \to [0,1]$ *by*

$$(h_1, \ldots, h_n)_l(x_1, y_1, \ldots, x_n, y_n) := \frac{1}{n} \sum_{i=1}^{n} l(h_i(x_i), y_i). \tag{12}$$

*We will also use* $\mathbf{h}_l$ *to denote* $(h_1, \ldots, h_n)_l$. *For any* $\mathcal{H}$ *in the hypothesis space family* $\mathbb{H}$, *define*

$$\mathcal{H}_l^n := \{(h_1, \ldots, h_n)_l \colon h_1, \ldots, h_n \in \mathcal{H}\}. \tag{13}$$

*Define*

$$\mathbb{H}_l^n := \bigcup_{\mathcal{H} \in \mathbb{H}} \mathcal{H}_l^n. \tag{14}$$

In the first part of the definition above, hypotheses $h\colon X \to Y$ are turned into functions $h_l$ mapping $X \times Y \to [0, 1]$ by composition with the loss function. $\mathcal{H}_l$ is then just the collection of all such functions where the original hypotheses come from $\mathcal{H}$. $\mathcal{H}_l$ is often called a *loss-function class*. In our case we are interested in the average loss across $n$ tasks, where each of the $n$ hypotheses is chosen from a fixed hypothesis space $\mathcal{H}$. This motivates the definition of $\mathbf{h}_l$ and $\mathcal{H}_l^n$. Finally, $\mathbb{H}_l^n$ is the collection of all $(h_1, \ldots, h_n)_l$, with the restriction that all $h_1, \ldots, h_n$ belong to a single hypothesis space $\mathcal{H} \in \mathbb{H}$.

**Definition 2.** *For each $\mathcal{H} \in \mathbb{H}$, define $\mathcal{H}^*\colon \mathcal{P} \to [0, 1]$ by*

$$\mathcal{H}^*(P) := \inf_{h \in \mathcal{H}} \mathrm{er}_P(h). \tag{15}$$

*For the hypothesis space family $\mathbb{H}$, define*

$$\mathbb{H}^* := \{\mathcal{H}^*\colon \mathcal{H} \in \mathbb{H}\}. \tag{16}$$

It is the "size" of $\mathbb{H}_l^n$ and $\mathbb{H}^*$ that controls how large the $(n, m)$-sample $\mathbf{z}$ must be to ensure $\hat{\mathrm{er}}_{\mathbf{z}}(\mathcal{H})$ and $\mathrm{er}_Q(\mathcal{H})$ are close uniformly over all $\mathcal{H} \in \mathbb{H}$. Their size will be defined in terms of certain covering numbers, and for this we need to define how to measure the distance between elements of $\mathbb{H}_l^n$ and also between elements of $\mathbb{H}^*$.

**Definition 3.** *Let $\mathbf{P} = (P_1, \ldots, P_n)$ be any sequence of $n$ probability distributions on $X \times Y$. For any $\mathbf{h}_l, \mathbf{h}_l' \in \mathbb{H}_l^n$, define*

$$d_{\mathbf{P}}(\mathbf{h}_l, \mathbf{h}_l') := \int_{(X \times Y)^n} |\mathbf{h}_l(x_1, y_1, \ldots, x_n, y_n) - \mathbf{h}_l'(x_1, y_1, \ldots, x_n, y_n)|$$
$$dP_1(x_1, y_1) \ldots dP_n(x_n, y_n) \tag{17}$$

*Similarly, for any distribution $Q$ on $\mathcal{P}$ and any $\mathcal{H}_1^*, \mathcal{H}_2^* \in \mathbb{H}^*$, define*

$$d_Q(\mathcal{H}_1^*, \mathcal{H}_2^*) := \int_{\mathcal{P}} |\mathcal{H}_1^*(P) - \mathcal{H}_2^*(P)| \ dQ(P) \tag{18}$$

It is easily verified that $d_{\mathbf{P}}$ and $d_Q$ are pseudo-metrics[4] on $\mathbb{H}_l^n$ and $\mathbb{H}^*$ respectively.

**Definition 4.** *An $\varepsilon$-cover of $(\mathbb{H}^*, d_Q)$ is a set $\{\mathcal{H}_1^*, \ldots, \mathcal{H}_N^*\}$ such that for all $\mathcal{H}^* \in \mathbb{H}^*$, $d_Q(\mathcal{H}^*, \mathcal{H}_i^*) \le \varepsilon$ for some $i = 1 \ldots N$. Note that we do not require the $\mathcal{H}_i^*$ to be contained in $\mathbb{H}^*$, just that they be measurable functions on $\mathcal{P}$. Let $\mathcal{N}(\varepsilon, \mathbb{H}^*, d_Q)$ denote the size of the smallest such cover. Define the* capacity *of $\mathbb{H}^*$ by*

$$\mathcal{C}(\varepsilon, \mathbb{H}^*) := \sup_Q \mathcal{N}(\varepsilon, \mathbb{H}^*, d_Q) \tag{19}$$

*where the supremum is over all probability measures on $\mathcal{P}$. $\mathcal{N}(\varepsilon, \mathbb{H}_l^n, d_{\mathbf{P}})$ is defined in a similar way, using $d_{\mathbf{P}}$ in place of $d_Q$. Define the* capacity *of $\mathbb{H}_l^n$ by:*

$$\mathcal{C}(\varepsilon, \mathbb{H}_l^n) := \sup_{\mathbf{P}} \mathcal{N}(\varepsilon, \mathbb{H}_l^n, d_{\mathbf{P}}) \tag{20}$$

*where now the supremum is over all sequences of $n$ probability measures on $X \times Y$.*

---

4. A pseudo-metric $d$ is a metric without the condition that $d(x, y) = 0 \Rightarrow x = y$.

### 2.4 Uniform Convergence for Bias Learners

Now we have enough machinery to state the main theorem. In the theorem the hypothesis space family is required to be *permissible*. Permissibility is discussed in detail in Appendix D, but note that it is a weak measure-theoretic condition satisfied by almost all "real-world" hypothesis space families. All logarithms are to base $e$.

**Theorem 2.** *Suppose $X$ and $Y$ are separable metric spaces and let $Q$ be any probability distribution on $\mathcal{P}$, the set of all distributions on $X \times Y$. Suppose $\mathbf{z}$ is an $(n, m)$-sample generated by sampling $n$ times from $\mathcal{P}$ according to $Q$ to give $P_1, \ldots, P_n$, and then sampling $m$ times from each $P_i$ to generate $z_i = \{(x_{i1}, y_{i1}), \ldots, (x_{im}, y_{im})\}, i = 1, \ldots, n$. Let $\mathbb{H} = \{\mathcal{H}\}$ be any permissible hypothesis space family. If the number of tasks $n$ satisfies*

$$ n \geq \max \left\{ \frac{256}{\varepsilon^2} \log \frac{8\mathcal{C}\left(\frac{\varepsilon}{32}, \mathbb{H}^*\right)}{\delta}, \frac{64}{\varepsilon^2} \right\}, \tag{21} $$

*and the number of examples $m$ of each task satisfies*

$$ m \geq \max \left\{ \frac{256}{n\varepsilon^2} \log \frac{8\mathcal{C}\left(\frac{\varepsilon}{32}, \mathbb{H}_l^n\right)}{\delta}, \frac{64}{\varepsilon^2} \right\}, \tag{22} $$

*then with probability at least $1 - \delta$ (over the $(n, m)$-sample $\mathbf{z}$), all $\mathcal{H} \in \mathbb{H}$ will satisfy*

$$ \mathrm{er}_Q(\mathcal{H}) \leq \mathrm{\acute{e}r}_{\mathbf{z}}(\mathcal{H}) + \varepsilon \tag{23} $$

*Proof.* See Appendix A. $\qquad\qquad\qquad\qquad\qquad\qquad\qquad\qquad\qquad\qquad\qquad\qquad\qquad\square$

There are several important points to note about Theorem 2:

1. Provided the capacities $\mathcal{C}(\varepsilon, \mathbb{H}^*)$ and $\mathcal{C}(\varepsilon, \mathbb{H}_l^n)$ are finite, the theorem shows that *any* bias learner that selects hypothesis spaces from $\mathbb{H}$ can bound its generalisation error $\mathrm{er}_Q(\mathcal{H})$ in terms of $\mathrm{\acute{e}r}_{\mathbf{z}}(\mathcal{H})$ for sufficiently large $(n, m)$-samples $\mathbf{z}$. Most bias learner's will not find the exact value of $\mathrm{\acute{e}r}_{\mathbf{z}}(\mathcal{H})$ because it involves finding the smallest error of any hypothesis $h \in \mathcal{H}$ on each of the $n$ training sets in $\mathbf{z}$. But any upper bound on $\mathrm{\acute{e}r}_{\mathbf{z}}(\mathcal{H})$ (found, for example by gradient descent on some error function) will still give an upper bound on $\mathrm{er}_Q(\mathcal{H})$. See Section 3.3.1 for a brief discussion on how this can be achieved in a feature learning setting.

2. In order to learn bias (in the sense that $\mathrm{er}_Q(\mathcal{H})$ and $\mathrm{\acute{e}r}_{\mathbf{z}}(\mathcal{H})$ are close uniformly over all $\mathcal{H} \in \mathbb{H}$), both the number of tasks $n$ and the number of examples of each task $m$ must be sufficiently large. This is intuitively reasonable because the bias learner must see both sufficiently many tasks to be confident of the nature of the environment, and sufficiently many examples of each task to be confident of the nature of each task.

3. Once the learner has found an $\mathcal{H} \in \mathbb{H}$ with a small value of $\mathrm{\acute{e}r}_{\mathbf{z}}(\mathcal{H})$, it can then use $\mathcal{H}$ to learn novel tasks $P$ drawn according to $Q$. One then has the following theorem bounding the sample complexity required for good generalisation when learning with $\mathcal{H}$ (the proof is very similar to the proof of the bound on $m$ in Theorem 2).

**Theorem 3.** *Let $z = \{(x_1, y_1), \ldots, (x_m, y_m)\}$ be a training set generated by sampling from $X \times Y$ according to some distribution $P$. Let $\mathcal{H}$ be a permissible hypothesis space. For all $\varepsilon, \delta$ with $0 < \varepsilon, \delta < 1$, if the number of training examples $m$ satisfies*

$$m \geq \max \left\{ \frac{64}{\varepsilon^2} \log \frac{4\mathcal{C}\left(\frac{\varepsilon}{16}, \mathcal{H}_l\right)}{\delta}, \frac{16}{\varepsilon^2} \right\} \tag{24}$$

*then with probability at least $1 - \delta$, all $h \in \mathcal{H}$ will satisfy*

$$\mathrm{er}_P(h) \leq \dot{\mathrm{er}}_z(h) + \varepsilon.$$

The capacity $\mathcal{C}(\varepsilon, \mathcal{H})$ appearing in equation (24) is defined in an analogous fashion to the capacities in Definition 4 (we just use the pseudo-metric $d_P(h_l, h_l') := \int_{X \times Y} |h_l(x, y) - h_l'(x, y)| \, dP(x, y)$). The important thing to note about Theorem 3 is that the number of examples required for good generalisation when learning novel tasks is proportional to the logarithm of the capacity of the learnt hypothesis space $\mathcal{H}$. In contrast, if the learner does not do any bias learning, it will have no reason to select one hypothesis space $\mathcal{H} \in \mathbb{H}$ over any other and consequently it would have to view as a candidate solution any hypothesis in any of the hypothesis spaces $\mathcal{H} \in \mathbb{H}$. Thus, its sample complexity will be proportional to the capacity of $\cup_{\mathcal{H} \in \mathbb{H}} \{\mathcal{H}_l\} = \mathbb{H}_l^1$, which in general will be considerably larger than the capacity of any individual $\mathcal{H} \in \mathbb{H}$. So by learning $\mathcal{H}$ the learner has *learnt to learn* in the environment $(\mathcal{P}, Q)$ in the sense that it needs far smaller training sets to learn novel tasks.

4. Having learnt a hypothesis space $\mathcal{H}$ with a small value of $\dot{\mathrm{er}}_{\mathbf{z}}(\mathcal{H})$, Theorem 2 tells us that with probability at least $1 - \delta$, the expected value of $\inf_{h \in \mathcal{H}} \mathrm{er}_P(h)$ on a novel task $P$ will be less than $\dot{\mathrm{er}}_{\mathbf{z}}(\mathcal{H}) + \varepsilon$. Of course, this does not rule out really bad performance on some tasks $P$. However, the probability of generating such "bad" tasks can be bounded. In particular, note that $\mathrm{er}_Q(\mathcal{H})$ is just the expected value of the function $\mathcal{H}^*$ over $\mathcal{P}$, and so by Markov's inequality, for $\gamma > 0$,

$$\Pr\left\{ P \colon \inf_{h \in \mathcal{H}} \mathrm{er}_P(h) \geq \gamma \right\} = \Pr\left\{ P \colon \mathcal{H}^*(P) \geq \gamma \right\}$$
$$\leq \frac{E_Q \mathcal{H}^*}{\gamma}$$
$$= \frac{\mathrm{er}_Q(\mathcal{H})}{\gamma}$$
$$\leq \frac{\dot{\mathrm{er}}_{\mathbf{z}}(\mathcal{H}) + \varepsilon}{\gamma} \quad \text{(with probability } 1 - \delta\text{)}.$$

5. Keeping the accuracy and confidence parameters $\varepsilon, \delta$ fixed, note that the number of examples required of each task for good generalisation obeys

$$m = O\left( \frac{1}{n} \log \mathcal{C}\left(\varepsilon, \mathbb{H}_l^n\right) \right). \tag{25}$$

So provided $\log \mathcal{C}(\varepsilon, \mathbb{H}_l^n)$ increases sublinearly with $n$, the upper bound on the number of examples required of each task will *decrease* as the number of tasks increases. This shows that for suitably constructed hypothesis space families it is possible to *share* information between tasks. This is discussed further after Theorem 4 below.

### 2.5 Choosing the Hypothesis Space Family $\mathbb{H}$.

Theorem 2 only provides conditions under which $\mathrm{\acute{e}r_z}(\mathcal{H})$ and $\mathrm{er}_Q(\mathcal{H})$ are close, it does not guarantee that $\mathrm{er}_Q(\mathcal{H})$ is actually small. This is governed by the choice of $\mathbb{H}$. If $\mathbb{H}$ contains a hypothesis space $\mathcal{H}$ with a small value of $\mathrm{er}_Q(\mathcal{H})$ and the learner is able to find an $\mathcal{H} \in \mathbb{H}$ minimizing error on the $(n, m)$ sample $\mathbf{z}$ (i.e., minimizing $\mathrm{\acute{e}r_z}(\mathcal{H})$), then, for sufficiently large $n$ and $m$, Theorem 2 ensures that with high probability $\mathrm{er}_Q(\mathcal{H})$ will be small. However, a bad choice of $\mathbb{H}$ will mean there is no hope of finding an $\mathcal{H}$ with small error. In this sense the choice of $\mathbb{H}$ represents the *hyper-bias* of the learner.

Note that from a sample complexity point of view, the *optimal* hypothesis space family to choose is one containing a single, minimal hypothesis space $\mathcal{H}$ that contains good solutions to all of the problems in the environment (or at least a set of problems with high $Q$-probability), and no more. For then there is no bias learning to do (because there is no choice to be made between hypothesis spaces), the output of the bias learning algorithm is guaranteed to be a good hypothesis space for the environment, and since the hypothesis space is minimal, learning any problem within the environment using $\mathcal{H}$ will require the smallest possible number of examples. However, this scenario is analogous to the trivial scenario in ordinary learning in which the learning algorithm contains a single, optimal hypothesis for the problem being learnt. In that case there is no learning to be done, just as there is no bias learning to be done if the correct hypothesis space is already known.

At the other extreme, if $\mathbb{H}$ contains a single hypothesis space $\mathcal{H}$ consisting of all possible functions from $X \to Y$ then bias learning is impossible because the bias learner cannot produce a restricted hypothesis space as output, and hence cannot produce a hypothesis space with improved sample complexity requirements on as yet unseen tasks.

Focussing on these two extremes highlights the minimal requirements on $\mathbb{H}$ for successful bias learning to occur: the hypothesis spaces $\mathcal{H} \in \mathbb{H}$ must be strictly smaller than the space of all functions $X \to Y$, but not so small or so "skewed" that none of them contain good solutions to a large majority of the problems in the environment.

It may seem that we have simply replaced the problem of selecting the right bias (i.e., selecting the right hypothesis space $\mathcal{H}$) with the equally difficult problem of selecting the right hyper-bias (i.e., the right hypothesis space family $\mathbb{H}$). However, in many cases selecting the right hyper-bias is far easier than selecting the right bias. For example, in Section 3 we will see how the feature selection problem may be viewed as a bias selection problem. Selecting the right features can be extremely difficult if one knows little about the environment, with intelligent trial-and-error typically the best one can do. However, in a bias learning scenario, one only has to specify that a set of features should exist, find a loosely parameterised set of features (for example neural networks), and then learn the features by sampling from multiple related tasks.

### 2.6 Learning Multiple Tasks

It may be that the learner is not interested in learning to learn, but just wants to learn a fixed set of $n$ tasks from the environment $(\mathcal{P}, Q)$. As in the previous section, we assume the learner starts out with a hypothesis space family $\mathbb{H}$, and also that it receives an $(n, m)$-sample $\mathbf{z}$ generated from the $n$ distributions $P_1, \ldots, P_n$. This time, however, the learner is simply looking for $n$ hypotheses $(h_1, \ldots, h_n)$, all contained in the same hypothesis space $\mathcal{H}$, such that the average generalization error of the $n$ hypotheses is minimal. Denoting $(h_1, \ldots, h_n)$ by $\mathbf{h}$ and writing $\mathbf{P} = (P_1, \ldots, P_n)$,

this error is given by:

$$\mathrm{er}_{\mathbf{P}}(\mathbf{h}) := \frac{1}{n} \sum_{i=1}^{n} \mathrm{er}_{P_i}(h_i) \tag{26}$$

$$= \frac{1}{n} \sum_{i=1}^{n} \int_{X \times Y} l(h_i(x), y) \, dP_i(x, y),$$

and the empirical loss of $\mathbf{h}$ on $\mathbf{z}$ is

$$\mathrm{\acute{e}r}_{\mathbf{z}}(\mathbf{h}) := \frac{1}{n} \sum_{i=1}^{n} \mathrm{\acute{e}r}_{z_i}(h_i) \tag{27}$$

$$= \frac{1}{n} \sum_{i=1}^{n} \frac{1}{m} \sum_{j=1}^{m} l(h_i(x_{ij}), y_{ij}).$$

As before, regardless of how the learner chooses $(h_1, \ldots, h_n)$, if we can prove a uniform bound on the probability of large deviation between $\mathrm{\acute{e}r}_{\mathbf{z}}(\mathbf{h})$ and $\mathrm{er}_{\mathbf{P}}(\mathbf{h})$ then any $(h_1, \ldots, h_n)$ that perform well on the training sets $\mathbf{z}$ will with high probability perform well on future examples of the same tasks.

**Theorem 4.** *Let $\mathbf{P} = (P_1, \ldots, P_n)$ be $n$ probability distributions on $X \times Y$ and let $\mathbf{z}$ be an $(n, m)$-sample generated by sampling $m$ times from $X \times Y$ according to each $P_i$. Let $\mathbb{H} = \{\mathcal{H}\}$ be any permissible hypothesis space family. If the number of examples $m$ of each task satisfies*

$$m \geq \max \left\{ \frac{64}{n\varepsilon^2} \log \frac{4\mathcal{C}(\frac{\varepsilon}{16}, \mathbb{H}_l^n)}{\delta}, \frac{16}{\varepsilon^2} \right\} \tag{28}$$

*then with probability at least $1 - \delta$ (over the choice of $\mathbf{z}$), any $\mathbf{h} \in \mathbb{H}^n$ will satisfy*

$$\mathrm{er}_{\mathbf{P}}(\mathbf{h}) \leq \mathrm{\acute{e}r}_{\mathbf{z}}(\mathbf{h}) + \varepsilon \tag{29}$$

*(recall Definition 4 for the meaning of $\mathcal{C}(\varepsilon, \mathbb{H}_l^n)$).*

*Proof.* Omitted (follow the proof of the bound on $m$ in Theorem 2). □

The bound on $m$ in Theorem 4 is virtually identical to the bound on $m$ in Theorem 2, and note again that it depends *inversely* on the number of tasks $n$ (assuming that the first part of the "max" expression is the dominate one). Whether this helps depends on the rate of growth of $\mathcal{C}(\frac{\varepsilon}{16}, \mathbb{H}_l^n)$ as a function of $n$. The following Lemma shows that this growth is always small enough to ensure that we never do worse by learning multiple tasks (at least in terms of the upper bound on the number of examples required per task).

**Lemma 5.** *For any hypothesis space family $\mathbb{H}$,*

$$\mathcal{C}\left(\varepsilon, \mathbb{H}_l^1\right) \leq \mathcal{C}\left(\varepsilon, \mathbb{H}_l^n\right) \leq \mathcal{C}\left(\varepsilon, \mathbb{H}_l^1\right)^n. \tag{30}$$

*Proof.* Let $K$ denote the set of all functions $(h_1, \ldots, h_n)_l$ where each $h_i$ can be a member of any hypothesis space $\mathcal{H} \in \mathbb{H}$ (recall Definition 1). Then $\mathbb{H}_l^n \subseteq K$ and so $\mathcal{C}(\varepsilon, \mathbb{H}_l^n) \leq \mathcal{C}(\varepsilon, K)$. By Lemma 29 in Appendix B, $\mathcal{C}(\varepsilon, K) \leq \mathcal{C}(\varepsilon, \mathbb{H}_l^1)^n$ and so the right hand inequality follows.

For the first inequality, let $P$ be any probability measure on $X \times Y$ and let $\mathbf{P}$ be the measure on $(X \times Y)^n$ obtained by using $P$ on the first copy of $X \times Y$ in the product, and ignoring all other elements of the product. Let $N$ be an $\varepsilon$-cover for $(\mathbb{H}_l^n, d_{\mathbf{P}})$. Pick any $h_l \in \mathbb{H}_l^1$ and let $(g_1, \ldots, g_n)_l \in N$ be such that $d_{\mathbf{P}}((h, h, \ldots, h)_l, (g_1, \ldots, g_n)_l) \leq \varepsilon$. But by construction, $d_{\mathbf{P}}((h, h, \ldots, h)_l, (g_1, \ldots, g_n)_l) = d_P(h, (g_1)_l)$, which establishes the first inequality. $\qquad\square$

By Lemma 5

$$\log \mathcal{C}(\varepsilon, \mathbb{H}_l^1) \leq \log \mathcal{C}(\varepsilon, \mathbb{H}_l^n) \leq n \log \mathcal{C}(\varepsilon, \mathbb{H}_l^1). \tag{31}$$

So keeping the accuracy parameters $\varepsilon$ and $\delta$ fixed, and plugging (31) into (28), we see that the upper bound on the number of examples required of each task never *increases* with the number of tasks, and at best decreases as $O(1/n)$. Although only an upper bound, this provides a strong hint that learning multiple related tasks should be advantageous on a "number of examples required per task" basis. In Section 3 it will be shown that for feature learning all types of behavior are possible, from no advantage at all to $O(1/n)$ decrease.

### 2.7 Dependence on $\varepsilon$

In Theorems 2, 3 and 4 the bounds on sample complexity all scale as $1/\varepsilon^2$. This behavior can be improved to $1/\varepsilon$ if the empirical loss is always guaranteed to be zero (i.e., we are in the realizable case). The same behavior results if we are interested in relative deviation between empirical and true loss, rather than absolute deviation. Formal theorems along these lines are stated in Appendix A.3.

## 3. Feature Learning

The use of restricted feature sets is nearly ubiquitous as a method of encoding bias in many areas of machine learning and statistics, including classification, regression and density estimation.

In this section we show how the problem of choosing a set of features for an environment of related tasks can be recast as a bias learning problem. Explicit bounds on $\mathcal{C}(\mathbb{H}^*, \varepsilon)$ and $\mathcal{C}(\mathbb{H}^n, \varepsilon)$ are calculated for general feature classes in Section 3.2. These bounds are applied to the problem of learning a neural network feature set in Section 3.3.

### 3.1 The Feature Learning Model

Consider the following quote from Vapnik (1996):

> The classical approach to estimating multidimensional functional dependencies is based on the following belief:
>
> Real-life problems are such that there exists a small number of "strong features," simple functions of which (say linear combinations) approximate well the unknown function. Therefore, it is necessary to carefully choose a low-dimensional feature space and then to use regular statistical techniques to construct an approximation.

In general a set of "strong features" may be viewed as a function $f\colon X \to V$ mapping the input space $X$ into some (typically lower) dimensional space $V$. Let $\mathcal{F} = \{f\}$ be a set of such feature maps (each $f$ may be viewed as a set of features $(f_1, \ldots, f_k)$ if $V = \mathbb{R}^k$). It is the $f$ that must be "carefully chosen" in the above quote. In general, the "simple functions of the features" may be represented as a class of functions $\mathcal{G}$ mapping $V$ to $Y$. If for each $f \in \mathcal{F}$ we define the hypothesis space $\mathcal{G} \circ f := \{g \circ f\colon g \in \mathcal{G}\}$, then we have the hypothesis space family $\mathbb{H}$

$$\mathbb{H} := \{\mathcal{G} \circ f\colon f \in \mathcal{F}\}. \tag{32}$$

Now the problem of "carefully choosing" the right features $f$ is equivalent to the bias learning problem "find the right hypothesis space $\mathcal{H} \in \mathbb{H}$". Hence, provided the learner is embedded within an environment of related tasks, and the capacities $\mathcal{C}(\mathbb{H}^*, \varepsilon)$ and $\mathcal{C}(\mathbb{H}_l^n, \varepsilon)$ are finite, Theorem 2 tells us that the feature set $f$ can be *learnt* rather than carefully chosen. This represents an important simplification, as choosing a set of features is often the most difficult part of any machine learning problem.

In Section 3.2 we give a theorem bounding $\mathcal{C}(\mathbb{H}^*, \varepsilon)$ and $\mathcal{C}(\mathbb{H}_l^n, \varepsilon)$ for general feature classes. The theorem is specialized to neural network classes in Section 3.3.

Note that we have forced the function class $\mathcal{G}$ to be the same for all feature maps $f$, although this is not necessary. Indeed variants of the results to follow can be obtained if $\mathcal{G}$ is allowed to vary with $f$.

### 3.2 Capacity Bounds for General Feature Classes

Notationally it is easier to view the feature maps $f$ as mapping from $X \times Y$ to $V \times Y$ by $(x, y) \mapsto (f(x), y)$, and also to absorb the loss function $l$ into the definition of $\mathcal{G}$ by viewing each $g \in \mathcal{G}$ as a map from $V \times Y$ into $[0, 1]$ via $(v, y) \mapsto l(g(v), y)$. Previously this latter function would have been denoted $g_l$ but in what follows we will drop the subscript $l$ where this does not cause confusion. The class to which $g_l$ belongs will still be denoted by $\mathcal{G}_l$.

With the above definitions let $\mathcal{G}_l \circ \mathcal{F} := \{g \circ f\colon g \in \mathcal{G}_l, f \in \mathcal{F}\}$. Define the capacity of $\mathcal{G}_l$ in the usual way,

$$\mathcal{C}(\varepsilon, \mathcal{G}_l) := \sup_P \mathcal{N}(\varepsilon, \mathcal{G}_l, d_P)$$

where the supremum is over all probability measures on $V \times Y$, and $d_P(g, g') := \int_{V \times Y} |g(v, y) - g'(v, y)| \, dP(v, y)$. To define the capacity of $\mathcal{F}$ we first define a pseudo-metric $d_{[P, \mathcal{G}_l]}$ on $\mathcal{F}$ by "pulling back" the $L^1$ metric on $\mathbb{R}$ through $\mathcal{G}_l$ as follows:

$$d_{[P, \mathcal{G}_l]}(f, f') := \int_{X \times Y} \sup_{g \in \mathcal{G}_l} |g \circ f(x, y) - g \circ f'(x, y)| \, dP(x, y). \tag{33}$$

It is easily verified that $d_{[P, \mathcal{G}_l]}$ is a pseudo-metric. Note that for $d_{[P, \mathcal{G}_l]}$ to be well defined the supremum over $\mathcal{G}_l$ in the integrand must be measurable. This is guaranteed if the hypothesis space family $\mathbb{H} = \{\mathcal{G}_l \circ f\colon f \in \mathcal{F}\}$ is permissible (Lemma 32, part 4). Now define $\mathcal{N}(\varepsilon, \mathcal{F}, d_{[P, \mathcal{G}_l]})$ to be the smallest $\varepsilon$-cover of the pseudo-metric space $(\mathcal{F}, d_{[P, \mathcal{G}_l]})$ and the $\varepsilon$-capacity of $\mathcal{F}$ (with respect to $\mathcal{G}_l$) as

$$\mathcal{C}_{\mathcal{G}_l}(\varepsilon, \mathcal{F}) := \sup_P \mathcal{N}(\varepsilon, \mathcal{F}, d_{[P, \mathcal{G}_l]})$$

where the supremum is over all probability measures on $X \times Y$. Now we can state the main theorem of this section.

**Theorem 6.** *Let $\mathbb{H}$ be a hypothesis space family as in equation* (32). *Then for all $\varepsilon, \varepsilon_1, \varepsilon_2 > 0$ with $\varepsilon = \varepsilon_1 + \varepsilon_2$,*

$$\mathcal{C}\left(\varepsilon, \mathbb{H}_l^n\right) \leq \mathcal{C}\left(\varepsilon_1, \mathcal{G}_l\right)^n \, \mathcal{C}_{\mathcal{G}_l}\left(\varepsilon_2, \mathcal{F}\right) \tag{34}$$

$$\mathcal{C}\left(\varepsilon, \mathbb{H}^*\right) \leq \mathcal{C}_{\mathcal{G}_l}\left(\varepsilon, \mathcal{F}\right) \tag{35}$$

*Proof.* See Appendix B. □

### 3.3 Learning Neural Network Features

In general, a set of features may be viewed as a map from the (typically high-dimensional) input space $\mathbb{R}^d$ to a much smaller dimensional space $\mathbb{R}^k$ ($k \ll d$). In this section we consider approximating such a feature map by a one-hidden-layer neural network with $d$ input nodes and $k$ output nodes (Figure 1). We denote the set of all such feature maps by $\{\Phi_w = (\phi_{w,1}, \ldots, \phi_{w,k}) \colon w \in D\}$ where $D$ is a bounded subset of $\mathbb{R}^W$ ($W$ is the number of weights (parameters) in the first two layers). This set is the $\mathcal{F}$ of the previous section.

Each feature $\phi_{w,i} \colon \mathbb{R}^d \to [0, 1]$, $i = 1, \ldots, k$ is defined by

$$\phi_{w,i}(x) := \sigma\left(\sum_{j=1}^{l} v_{ij} h_j(x) + v_{il+1}\right) \tag{36}$$

where $h_j(x)$ is the output of the $j$th node in the first hidden layer, $(v_{i1}, \ldots, v_{il+1})$ are the output node parameters for the $i$th feature and $\sigma$ is a "sigmoid" squashing function $\sigma \colon \mathbb{R} \to [0, 1]$. Each first layer hidden node $h_i \colon \mathbb{R}^d \to \mathbb{R}$, $i = 1, \ldots, l$, computes

$$h_i(x) := \sigma\left(\sum_{j=1}^{d} u_{ij} x_j + u_{id+1}\right) \tag{37}$$

where $(u_{i1}, \ldots, u_{id+1})$ are the hidden node's parameters. We assume $\sigma$ is Lipschitz.[5] The weight vector for the entire feature map is thus

$$w = (u_{11}, \ldots, u_{1d+1}, \ldots, u_{l1}, \ldots, u_{ld+1}, v_{11}, \ldots, v_{1l+1}, \ldots, v_{k1}, \ldots, v_{kl+1})$$

and the total number of feature parameters $W = l(d + 1) + k(l + 1)$.

For argument's sake, assume the "simple functions" of the features (the class $\mathcal{G}$ of the previous section) are squashed affine maps using the same sigmoid function $\sigma$ above (in keeping with the "neural network" flavor of the features). Thus, each setting of the feature weights $w$ generates a hypothesis space:

$$\mathcal{H}_w := \left\{\sigma\left(\sum_{i=1}^{k} \alpha_i \phi_{w,i} + \alpha_{k+1}\right) : (\alpha_1, \ldots, \alpha_{k+1}) \in D'\right\}, \tag{38}$$

where $D'$ is a bounded subset of $\mathbb{R}^{k+1}$. The set of all such hypothesis spaces,

$$\mathbb{H} := \{\mathcal{H}_w \colon w \in D\} \tag{39}$$

---

5. $\sigma$ is Lipschitz if there exists a constant $K$ such that $|\sigma(x) - \sigma(x')| \leq K|x - x'|$ for all $x, x' \in \mathbb{R}$.

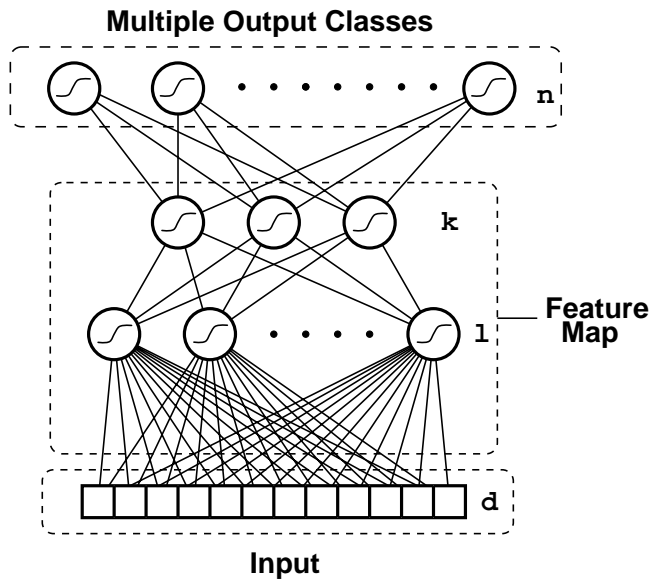

Figure 1: Neural network for feature learning. The feature map is implemented by the first two hidden layers. The $n$ output nodes correspond to the $n$ different tasks in the $(n, m)$-sample $\mathbf{z}$. Each node in the network computes a squashed linear function of the nodes in the previous layer.

is a hypothesis space family. The restrictions on the output layer weights $(\alpha_1, \ldots, \alpha_{k+1})$ and feature weights $w$, and the restriction to a Lipschitz squashing function are needed to obtain finite upper bounds on the covering numbers in Theorem 2.

Finding a good set of features for the environment $(\mathcal{P}, Q)$ is equivalent to finding a good hypothesis space $\mathcal{H}_w \in \mathbb{H}$, which in turn means finding a good set of feature map parameters $w$.

As in Theorem 2, the correct set of features may be learnt by finding a hypothesis space with small error on a sufficiently large $(n, m)$-sample $\mathbf{z}$. Specializing to squared loss, in the present framework the empirical loss of $\mathcal{H}_w$ on $\mathbf{z}$ (equation (8)) is given by

$$\hat{\text{er}}_{\mathbf{z}}(\mathcal{H}_w) = \frac{1}{n} \sum_{i=1}^{n} \inf_{(\alpha_0, \alpha_1, \ldots, \alpha_k) \in D'} \frac{1}{m} \sum_{j=1}^{m} \left[ \sigma \left( \sum_{l=1}^{k} \alpha_l \phi_{w,l}(x_{ij}) + \alpha_0 \right) - y_{ij} \right]^2 \quad (40)$$

Since our sigmoid function $\sigma$ only has range $[0, 1]$, we also restrict the outputs $Y$ to this range.

### 3.3.1 ALGORITHMS FOR FINDING A GOOD SET OF FEATURES

Provided the squashing function $\sigma$ is differentiable, gradient descent (with a small variation on backpropagation to compute the derivatives) can be used to find feature weights $w$ minimizing (40) (or at least a local minimum of (40)). The only extra difficulty over and above ordinary gradient descent is the appearance of "inf" in the definition of $\hat{\text{er}}_{\mathbf{z}}(\mathcal{H}_w)$. The solution is to perform gradient descent over both the output parameters $(\alpha_0, \ldots, \alpha_k)$ for each node and the feature weights $w$. For more details see Baxter (1995b) and Baxter (1995a, chapter 4), where empirical results supporting the theoretical results presented here are also given.

### 3.3.2 Sample Complexity Bounds for Neural-Network Feature Learning

The size of $\mathbf{z}$ ensuring that the resulting features will be good for learning novel tasks from the same environment is given by Theorem 2. All we have to do is compute the logarithm of the covering numbers $\mathcal{C}(\varepsilon, \mathbb{H}_l^n)$ and $\mathcal{C}(\varepsilon, \mathbb{H}^*)$.

**Theorem 7.** *Let* $\mathbb{H} = \left\{ \mathcal{H}_w : w \in \mathbb{R}^W \right\}$ *be a hypothesis space family where each* $\mathcal{H}_w$ *is of the form*

$$\mathcal{H}_w := \left\{ \sigma \left( \sum_{i=1}^k \alpha_i \phi_{w,i}(\cdot) + \alpha_0 \right) : (\alpha_1, \ldots, \alpha_k) \in \mathbb{R}^k \right\},$$

*where* $\Phi_w = (\phi_{w,1}, \ldots, \phi_{w,k})$ *is a neural network with* $W$ *weights mapping from* $\mathbb{R}^d$ *to* $\mathbb{R}^k$. *If the feature weights* $w$ *and the output weights* $\alpha_0, \alpha_1, \ldots, \alpha_k$ *are bounded, the squashing function* $\sigma$ *is Lipschitz,* $l$ *is squared loss, and the output space* $Y = [0, 1]$ *(any bounded subset of* $\mathbb{R}$ *will do), then there exist constants* $\kappa, \kappa'$ *(independent of* $\varepsilon$, $W$ *and* $k$*) such that for all* $\varepsilon > 0$,

$$\log \mathcal{C}(\varepsilon, \mathbb{H}_l^n) \leq 2\left((k+1)n + W\right)\log \frac{\kappa}{\varepsilon} \tag{41}$$

$$\log \mathcal{C}(\varepsilon, \mathbb{H}^*) \leq 2W \log \frac{\kappa'}{\varepsilon} \tag{42}$$

*(recall that we have specialized to squared loss here).*

*Proof.* See Appendix B. $\qquad\qquad\qquad\qquad\qquad\qquad\qquad\qquad\qquad\qquad\qquad$ $\square$

Noting that our neural network hypothesis space family $\mathbb{H}$ is permissible, plugging (41) and (42) into Theorem 2 gives the following theorem.

**Theorem 8.** *Let* $\mathbb{H} = \{\mathcal{H}_w\}$ *be a hypothesis space family where each hypothesis space* $\mathcal{H}_w$ *is a set of squashed linear maps composed with a neural network feature map, as above. Suppose the number of features is* $k$, *and the total number of feature weights is W. Assume all feature weights and output weights are bounded, and the squashing function* $\sigma$ *is Lipschitz. Let* $\mathbf{z}$ *be an* $(n, m)$-*sample generated from the environment* $(\mathcal{P}, Q)$. *If*

$$n \geq O\left(\frac{1}{\varepsilon^2}\left[W \log \frac{1}{\varepsilon} + \log \frac{1}{\delta}\right]\right), \tag{43}$$

*and*

$$m \geq O\left(\frac{1}{\varepsilon^2}\left[\left(k + 1 + \frac{W}{n}\right)\log \frac{1}{\varepsilon} + \frac{1}{n}\log \frac{1}{\delta}\right]\right) \tag{44}$$

*then with probability at least* $1 - \delta$ *any* $\mathcal{H}_w \in \mathbb{H}$ *will satisfy*

$$\mathrm{er}_Q(\mathcal{H}_w) \leq \hat{\mathrm{er}}_{\mathbf{z}}(\mathcal{H}_w) + \varepsilon. \tag{45}$$

### 3.3.3 DISCUSSION

1. Keeping the accuracy and confidence parameters $\varepsilon$ and $\delta$ fixed, the upper bound on the number of examples required of each task behaves like $O(k + W/n)$. If the learner is simply learning $n$ fixed tasks (rather than learning to learn), then the same upper bound also applies (recall Theorem 4).

2. Note that if we do away with the feature map altogether then $W = 0$ and the upper bound on $m$ becomes $O(k)$, independent of $n$ (apart from the less important $\delta$ term). So in terms of the upper bound, learning $n$ tasks becomes just as hard as learning one task. At the other extreme, if we fix the output weights then effectively $k = 0$ and the number of examples required of each task decreases as $O(W/n)$. Thus a range of behavior in the number of examples required of each task is possible: from no improvement at all to an $O(1/n)$ decrease as the number of tasks $n$ increases (recall the discussion at the end of Section 2.6).

3. Once the feature map is learnt (which can be achieved using the techniques outlined in Baxter, 1995b; Baxter & Bartlett, 1998; Baxter, 1995a, chapter 4), only the output weights have to be estimated to learn a novel task. Again keeping the accuracy parameters fixed, this requires no more that $O(k)$ examples. Thus, as the number of tasks learnt increases, the upper bound on the number of examples required of each task decays to the minimum possible, $O(k)$.

4. If the "small number of strong features" assumption is correct, then $k$ will be small. However, typically we will have very little idea of what the features are, so to be confident that the neural network is capable of implementing a good feature set it will need to be very large, implying $W \gg k$. $O(k + W/n)$ decreases most rapidly with increasing $n$ when $W \gg k$, so at least in terms of the upper bound on the number of examples required per task, learning small feature sets is an ideal application for bias learning. However, the upper bound on the number of tasks does not fare so well as it scales as $O(W)$.

### 3.3.4 COMPARISON WITH TRADITIONAL MULTIPLE-CLASS CLASSIFICATION

A special case of this multi-task framework is one in which the marginal distribution on the input space $P_{i|X}$ is the same for each task $i = 1, \ldots, n$, and all that varies between tasks is the conditional distribution over the output space $Y$. An example would be a multi-class problem such as face recognition, in which $Y = \{1, \ldots, n\}$ where $n$ is the number of faces to be recognized and the marginal distribution on $X$ is simply the "natural" distribution over images of those faces. In that case, if for every example $x_{ij}$ we have—in addition to the sample $y_{ij}$ from the $i$th task's conditional distribution on $Y$—samples from the remaining $n - 1$ conditional distributions on $Y$, then we can view the $n$ training sets containing $m$ examples each as one large training set for the multi-class problem with $mn$ examples altogether. The bound on $m$ in Theorem 8 states that $mn$ should be $O(nk + W)$, or proportional to the total number of parameters in the network, a result we would expect from[6] (Haussler, 1992).

So when specialized to the traditional multiple-class, single task framework, Theorem 8 is consistent with the bounds already known. However, as we have already argued, problems such as face recognition are not really single-task, multiple-class problems. They are more appropriately viewed as a (potentially infinite) collection of distinct binary classification problems. In that case, the goal of bias learning is not to find a single $n$-output network that can classify some subset of $n$ faces well. It is to learn a set of features that can reliably be used as a fixed preprocessing for distinguishing any single face from other faces. This is the new thing provided by Theorem 8: it tells us that provided we have trained our $n$-output neural network on sufficiently many examples of *sufficiently many tasks*, we can be confident that the common feature map learnt for those $n$ tasks will be good for learning *any* new, as yet unseen task, provided the new task is drawn from the same distribution that generated the training tasks. In addition, learning the new task only requires estimating the $k$ output node parameters for that task, a vastly easier problem than estimating the parameters of the entire network, from both a sample and computational complexity perspective. Also, since we have high confidence that the learnt features will be good for learning novel tasks drawn from the same environment, those features are themselves a candidate for further study to learn more about the nature of the environment. The same claim could not be made if the features had been learnt on too small a set of tasks to guarantee generalization to novel tasks, for then it is likely that the features would implement idiosyncrasies specific to those tasks, rather than "invariances" that apply across all tasks.

---

6. If each example can be classified with a "large margin" then naive parameter counting can be improved upon (Bartlett, 1998).

When viewed from a bias (or feature) learning perspective, rather than a traditional $n$-class classification perspective, the bound $m$ on the number of examples required of each task takes on a somewhat different meaning. It tells us that provided $n$ is large (i.e., we are collecting examples of a large number tasks), then we really only need to collect a few more examples than we would otherwise have to collect if the feature map was already known ($k + W/n$ examples vs. $k$ examples). So it tells us that the burden imposed by feature learning can be made negligibly small, at least when viewed from the perspective of the sampling burden required of each task.

## 3.4 Learning Multiple Tasks with Boolean Feature Maps

Ignoring the accuracy and confidence parameters $\varepsilon$ and $\delta$, Theorem 8 shows that the number of examples required of each task when learning $n$ tasks with a common neural-network feature map is bounded above by $O(k + W/n)$, where $k$ is the number of features and $W$ is the number of adjustable parameters in the feature map. Since $O(k)$ examples are required to learn a single task once the true features are known, this shows that the upper bound on the number of examples required of each task decays (in order) to the minimum possible as the number of tasks $n$ increases. This suggests that learning multiple tasks is advantageous, but to be truly convincing we need to prove a lower bound of the same form. Proving lower bounds in a real-valued setting ($Y = \mathbb{R}$) is complicated by the fact that a single example can convey an infinite amount of information, so one typically has to make extra assumptions, such as that the targets $y \in Y$ are corrupted by a noise process. Rather than concern ourselves with such complications, in this section we restrict our attention to Boolean hypothesis space families (meaning each hypothesis $h \in \mathbb{H}^1$ maps to $Y = \{\pm 1\}$ and we measure error by discrete loss $l(h(x), y) = 1$ if $h(x) \neq y$ and $l(h(x), y) = 0$ otherwise).

We show that the sample complexity for learning $n$ tasks with a Boolean hypothesis space family $\mathbb{H}$ is controlled by a "VC dimension" type parameter $d_{\mathbb{H}}(n)$ (that is, we give nearly matching upper and lower bounds involving $d_{\mathbb{H}}(n)$). We then derive bounds on $d_{\mathbb{H}}(n)$ for the hypothesis space family considered in the previous section with the Lipschitz sigmoid function $\sigma$ replaced by a hard threshold (linear threshold networks).

As well as the bound on the number of examples required per task for good generalization across those tasks, Theorem 8 also shows that features performing well on $O(W)$ *tasks* will generalize well to novel tasks, where $W$ is the number of parameters in the feature map. Given that for many feature learning problems $W$ is likely to be quite large (recall Note 4 in Section 3.3.3), it would be useful to know that $O(W)$ tasks are in fact *necessary* without further restrictions on the environmental distributions $Q$ generating the tasks. Unfortunately, we have not yet been able to show such a lower bound.

There is some empirical evidence suggesting that in practice the upper bound on the number of tasks may be very weak. For example, in Baxter and Bartlett (1998) we reported experiments in which a set of neural network features learnt on a subset of only 400 Japanese characters turned out to be good enough for classifying some 2600 unseen characters, even though the features contained several hundred thousand parameters. Similar results may be found in Intrator and Edelman (1996) and in the experiments reported in Thrun (1996) and Thrun and Pratt (1997, chapter 8). While this gap between experiment and theory may be just another example of the looseness inherent in general bounds, it may also be that the analysis can be tightened. In particular, the bound on the number of tasks is insensitive to the size of the class of output functions (the class $\mathcal{G}$ in Section 3.1), which may be where the looseness has arisen.

### 3.4.1 UPPER AND LOWER BOUNDS FOR LEARNING **n** TASKS WITH BOOLEAN HYPOTHESIS SPACE FAMILIES

First we recall some concepts from the theory of Boolean function learning. Let $\mathcal{H}$ be a class of Boolean functions on $X$ and $x = (x_1, \ldots, x_m) \in X^m$. $\mathcal{H}_{|x}$ is the set of all binary vectors obtainable by applying functions in $\mathcal{H}$ to $x$:

$$\mathcal{H}_{|x} := \{(h(x_1), \ldots, h(x_m)) \colon h \in \mathcal{H}\}.$$

Clearly $|\mathcal{H}_{|x}| \leq 2^m$. If $|\mathcal{H}_{|x}| = 2^m$ we say $\mathcal{H}$ *shatters* $x$. The *growth function* of $\mathcal{H}$ is defined by

$$\Pi_{\mathcal{H}}(m) := \max_{x \in X^m} |\mathcal{H}_{|x}|.$$

The *Vapnik-Chervonenkis dimension* $\mathrm{VCdim}(\mathcal{H})$ is the size of the largest set shattered by $\mathcal{H}$:

$$\mathrm{VCdim}(\mathcal{H}) := \max\{m : \Pi_{\mathcal{H}}(m) = 2^m\}.$$

An important result in the theory of learning Boolean functions is Sauer's Lemma (Sauer, 1972), of which we will also make use.

**Lemma 9 (Sauer's Lemma).** *For a Boolean function class $\mathcal{H}$ with $\mathrm{VCdim}(\mathcal{H}) = d$,*

$$\Pi_{\mathcal{H}}(m) \leq \sum_{i=0}^{d} \binom{m}{i} \leq \left(\frac{em}{d}\right)^d,$$

*for all positive integers $m$.*

We now generalize these concepts to learning $n$ tasks with a Boolean hypothesis space family.

**Definition 5.** *Let $\mathbb{H}$ be a Boolean hypothesis space family. Denote the $n \times m$ matrices over the input space $X$ by $X^{(n,m)}$. For each $\mathbf{x} \in X^{(n,m)}$ and $\mathcal{H} \in \mathbb{H}$, define $\mathcal{H}_{|\mathbf{x}}$ to be the set of (binary) matrices,*

$$\mathcal{H}_{|\mathbf{x}} := \left\{ \begin{bmatrix} h_1(x_{11}) & \cdots & h_1(x_{1m}) \\ \vdots & \ddots & \vdots \\ h_n(x_{n1}) & \cdots & h_n(x_{nm}) \end{bmatrix} : h_1, \ldots, h_n \in \mathcal{H} \right\}.$$

*Define*

$$\mathbb{H}_{|\mathbf{x}} := \bigcup_{\mathcal{H} \in \mathbb{H}} \mathcal{H}_{|\mathbf{x}}.$$

*Now for each $n > 0, m > 0$, define $\Pi_{\mathbb{H}}(n,m)$ by*

$$\Pi_{\mathbb{H}}(n,m) := \max_{\mathbf{x} \in X^{(n,m)}} \left| \mathbb{H}_{|\mathbf{x}} \right|.$$

*Note that $\Pi_{\mathbb{H}}(n,m) \leq 2^{nm}$. If $\left| \mathbb{H}_{|\mathbf{x}} \right| = 2^{nm}$ we say $\mathbb{H}$ **shatters** the matrix $\mathbf{x}$. For each $n > 0$ let*

$$d_{\mathbb{H}}(n) := \max\{m \colon \Pi_{\mathbb{H}}(n,m) = 2^{nm}\}.$$

*Define*

$$\overline{d}(\mathbb{H}) := \mathrm{VCdim}(\mathbb{H}^1) \quad and$$
$$\underline{d}(\mathbb{H}) := \max_{\mathcal{H} \in \mathbb{H}} \mathrm{VCdim}(\mathcal{H}).$$

**Lemma 10.**

$$\overline{d}(\mathbb{H}) \geq \underline{d}(\mathbb{H})$$
$$d_{\mathbb{H}}(n) \geq \max\left\{ \left\lfloor \frac{\overline{d}(\mathbb{H})}{n} \right\rfloor, \underline{d}(\mathbb{H}) \right\} \geq \frac{1}{2}\left( \left\lfloor \frac{\overline{d}(\mathbb{H})}{n} \right\rfloor + \underline{d}(\mathbb{H}) \right)$$

*Proof.* The first inequality is trivial from the definitions. To get the second term in the maximum in the second inequality, choose an $\mathcal{H} \in \mathbb{H}$ with $\mathrm{VCdim}(\mathcal{H}) = \underline{d}(\mathbb{H})$ and construct a matrix $\mathbf{x} \in X^{(n,m)}$ whose rows are of length $\underline{d}(\mathbb{H})$ and are shattered by $\mathcal{H}$. Then clearly $\mathbb{H}$ shatters $\mathbf{x}$. For the first term in the maximum take a sequence $x = (x_1, \ldots, x_{\overline{d}(\mathbb{H})})$ shattered by $\mathbb{H}^1$ (the hypothesis space consisting of the union over all hypothesis spaces from $\mathbb{H}$), and distribute its elements equally among the rows of $\mathbf{x}$ (throw away any leftovers). The set of matrices

$$\left\{ \begin{bmatrix} h(x_{11}) & \cdots & h(x_{1m}) \\ \vdots & \ddots & \vdots \\ h(x_{n1}) & \cdots & h(x_{nm}) \end{bmatrix} : h \in \mathbb{H}^1 \right\}.$$

where $m = \lfloor \overline{d}(\mathbb{H})/n \rfloor$ is a subset of $\mathbb{H}_{|\mathbf{x}}$ and has size $2^{nm}$. $\square$

**Lemma 11.**

$$\Pi_{\mathbb{H}}(n,m) \leq \left[ \frac{em}{d_{\mathbb{H}}(n)} \right]^{nd_{\mathbb{H}}(n)}$$

*Proof.* Observe that for each $n$, $\Pi_{\mathbb{H}}(n, m) = \Pi_{\mathcal{H}}(nm)$ where $\mathcal{H}$ is the collection of all Boolean functions on sequences $x_1, \ldots, x_{nm}$ obtained by first choosing $n$ functions $h_1, \ldots, h_n$ from some $\mathcal{H} \in \mathbb{H}$, and then applying $h_1$ to the first $m$ examples, $h_2$ to the second $m$ examples and so on. By the definition of $d_{\mathbb{H}}(n)$, $\mathrm{VCdim}(\mathcal{H}) = n d_{\mathbb{H}}(n)$, hence the result follows from Lemma 9 applied to $\mathcal{H}$. □

If one follows the proof of Theorem 4 (in particular the proof of Theorem 18 in Appendix A) then it is clear that for all $\epsilon > 0$, $\mathcal{C}(\mathbb{H}_l^n, \varepsilon)$ may be replaced by $\Pi_{\mathbb{H}}(n, 2m)$ in the Boolean case. Making this replacement in Theorem 18, and using the choices of $\alpha, \nu$ from the discussion following Theorem 26, we obtain the following bound on the probability of large deviation between empirical and true performance in this Boolean setting.

**Theorem 12.** *Let $\mathbf{P} = (P_1, \ldots, P_n)$ be $n$ probability distributions on $X \times \{\pm 1\}$ and let $\mathbf{z}$ be an $(n, m)$-sample generated by sampling $m$ times from $X \times \{\pm 1\}$ according to each $P_i$. Let $\mathbb{H} = \{\mathcal{H}\}$ be any permissible Boolean hypothesis space family. For all $0 < \epsilon \leq 1$,*

$$\Pr\{\mathbf{z}: \exists \mathbf{h} \in \mathbb{H}^n: \mathrm{er}_{\mathbf{P}}(\mathbf{h}) \geq \hat{\mathrm{er}}_{\mathbf{z}}(\mathbf{h}) + \varepsilon\} \leq 4\Pi_{\mathbb{H}}(n, 2m) \exp(-\epsilon^2 nm/64). \quad (46)$$

**Corollary 13.** *Under the conditions of Theorem 12, if the number of examples $m$ of each task satisfies*

$$m \geq \frac{88}{\varepsilon^2}\left[2d_{\mathbb{H}}(n)\log\frac{22}{\varepsilon} + \frac{1}{n}\log\frac{4}{\delta}\right] \quad (47)$$

*then with probability at least $1 - \delta$ (over the choice of $\mathbf{z}$), any $\mathbf{h} \in \mathbb{H}^n$ will satisfy*

$$\mathrm{er}_{\mathbf{P}}(\mathbf{h}) \leq \hat{\mathrm{er}}_{\mathbf{z}}(\mathbf{h}) + \varepsilon \quad (48)$$

*Proof.* Applying Theorem 12, we require

$$4\Pi_{\mathbb{H}}(n, 2m) \exp(-\epsilon^2 nm/64) \leq \delta,$$

which is satisfied if

$$m \geq \frac{64}{\epsilon^2}\left[d_{\mathbb{H}}(n)\log\frac{2em}{d_{\mathbb{H}}(n)} + \frac{1}{n}\log\frac{4}{\delta}\right], \quad (49)$$

where we have used Lemma 11. Now, for all $a \geq 1$, if

$$m = \left(1 + \frac{1}{e}\right) a \log\left(1 + \frac{1}{e}\right) a,$$

then $m \geq a \log m$. So setting $a = 64 d_{\mathbb{H}}(n)/\varepsilon^2$, (49) is satisfied if

$$m \geq \frac{88}{\varepsilon^2}\left[2d_{\mathbb{H}}(n)\log\frac{22}{\varepsilon} + \frac{1}{n}\log\frac{4}{\delta}\right].$$

□

Corollary 13 shows that any algorithm learning $n$ tasks using the hypothesis space family $\mathbb{H}$ requires no more than

$$m = O\left(\frac{1}{\varepsilon^2}\left[d_{\mathbb{H}}(n)\log\frac{1}{\varepsilon} + \frac{1}{n}\log\frac{1}{\delta}\right]\right) \tag{50}$$

examples of each task to ensure that with high probability the average true error of any $n$ hypotheses it selects from $\mathbb{H}^n$ is within $\varepsilon$ of their average empirical error on the sample $\mathbf{z}$. We now give a theorem showing that if the learning algorithm is required to produce $n$ hypotheses whose average true error is within $\varepsilon$ of the *best possible error* (achievable using $\mathbb{H}^n$) for an arbitrary sequence of distributions $P_1, \ldots, P_n$, then within a $\log\frac{1}{\varepsilon}$ factor the number of examples in equation (50) is also necessary.

For any sequence $\mathbf{P} = (P_1, \ldots, P_n)$ of $n$ probability distributions on $X \times \{\pm 1\}$, define $\mathrm{opt}_{\mathbf{P}}(\mathbb{H}^n)$ by

$$\mathrm{opt}_{\mathbf{P}}(\mathbb{H}^n) := \inf_{\mathbf{h} \in \mathbb{H}^n} \mathrm{er}_{\mathbf{P}}(\mathbf{h}).$$

**Theorem 14.** *Let $\mathbb{H}$ be a Boolean hypothesis space family such that $\mathbb{H}^1$ contains at least two functions. For each $n = 1, 2, \ldots$, let $\mathcal{A}_n$ be any learning algorithm taking as input $(n, m)$-samples $\mathbf{z} \in (X \times \{\pm 1\})^{(n,m)}$ and producing as output $n$ hypotheses $\mathbf{h} = (h_1, \ldots, h_n) \in \mathbb{H}^n$. For all $0 < \varepsilon < 1/64$ and $0 < \delta < 1/64$, if*

$$m < \frac{1}{\varepsilon^2}\left[\frac{d_{\mathbb{H}}(n)}{616} + (1 - \varepsilon^2)\frac{1}{n}\log\left(\frac{1}{8\delta(1-2\delta)}\right)\right]$$

*then there exist distributions $\mathbf{P} = (P_1, \ldots, P_n)$ such that with probability at least $\delta$ (over the random choice of $\mathbf{z}$),*

$$\mathrm{er}_{\mathbf{P}}(\mathcal{A}_n(\mathbf{z})) > \mathrm{opt}_{\mathbf{P}}(\mathbb{H}^n) + \varepsilon$$

*Proof.* See Appendix C $\qquad\qquad\qquad\qquad\qquad\qquad\qquad\qquad\qquad\qquad\qquad\qquad\qquad\qquad\Box$

### 3.4.2 Linear Threshold Networks

Theorems 13 and 14 show that within constants and a $\log(1/\varepsilon)$ factor, the sample complexity of learning $n$ tasks using the Boolean hypothesis space family $\mathbb{H}$ is controlled by the complexity parameter $d_{\mathbb{H}}(n)$. In this section we derive bounds on $d_{\mathbb{H}}(n)$ for hypothesis space families constructed as thresholded linear combinations of Boolean feature maps. Specifically, we assume $\mathbb{H}$ is of the form given by (39), (38), (37) and (36), where now the squashing function $\sigma$ is replaced with a hard threshold:

$$\sigma(x) := \begin{cases} 1 & \text{if } x \geq 0, \\ -1 & \text{otherwise,} \end{cases}$$

and we don't restrict the range of the feature and output layer weights. Note that in this case the proof of Theorem 8 does not carry through because the constants $\kappa, \kappa'$ in Theorem 7 depend on the Lipschitz bound on $\sigma$.

**Theorem 15.** *Let $\mathbb{H}$ be a hypothesis space family of the form given in (39), (38), (37) and (36), with a hard threshold sigmoid function $\sigma$. Recall that the parameters $d$, $l$ and $k$ are the input dimension, number of hidden nodes in the feature map and number of features (output nodes in the feature map)* *respectively. Let $W := l(d + 1) + k(l + 1)$ (the number of adjustable parameters in the feature map). Then,*

$$d_{\mathbb{H}}(n) \leq 2 \left( \frac{W}{n} + k + 1 \right) \log_2 \left( 2e(k + l + 1) \right).$$

*Proof.* Recall that for each $w \in \mathbb{R}^W$, $\Phi_w \colon \mathbb{R}^d \to \mathbb{R}^k$ denotes the feature map with parameters $w$. For each $\mathbf{x} \in X^{(n,m)}$, let $\Phi_{w|\mathbf{x}}$ denote the matrix

$$\left[ \begin{array}{ccc} \Phi_w(x_{11}) & \cdots & \Phi_w(x_{1m}) \\ \vdots & \ddots & \vdots \\ \Phi_w(x_{n1}) & \cdots & \Phi_w(x_{nm}) \end{array} \right].$$

Note that $\mathbb{H}_{|\mathbf{x}}$ is the set of all binary $n \times m$ matrices obtainable by composing thresholded linear functions with the elements of $\Phi_{w|\mathbf{x}}$, with the restriction that the same function must be applied to each element in a row (but the functions may differ between rows). With a slight abuse of notation, define

$$\Pi_{\Phi}(n, m) := \max_{\mathbf{x} \in X^{(n,m)}} \left| \left\{ \Phi_{w|\mathbf{x}} \colon w \in \mathbb{R}^W \right\} \right|.$$

Fix $\mathbf{x} \in X^{(n,m)}$. By Sauer's Lemma, each node in the first hidden layer of the feature map computes at most $(emn/(d + 1))^{d+1}$ functions on the $nm$ input vectors in $\mathbf{x}$. Thus, there can be at most $(emn/(d + 1))^{l(d+1)}$ distinct functions from the input to the output of the first hidden layer on the $nm$ points in $\mathbf{x}$. Fixing the first hidden layer parameters, each node in the second layer of the feature map computes at most $(emn/(l + 1))^{l+1}$ functions on the image of $\mathbf{x}$ produced at the output of the first hidden layer. Thus the second hidden layer computes no more than $(emn/(l + 1))^{k(l+1)}$ functions on the output of the first hidden layer on the $nm$ points in $\mathbf{x}$. So, in total,

$$\Pi_{\Phi}(n, m) \leq \left( \frac{emn}{d + 1} \right)^{l(d+1)} \left( \frac{emn}{l + 1} \right)^{k(l+1)}.$$

Now, for each possible matrix $\Phi_{w|\mathbf{x}}$, the number of functions computable on each row of $\Phi_{w|\mathbf{x}}$ by a thresholded linear combination of the output of the feature map is at most $(em/(k + 1))^{k+1}$. Hence, the number of binary sign assignments obtainable by applying linear threshold functions to all the rows is at most $(em/(k + 1))^{n(k+1)}$. Thus,

$$\Pi_{\mathbb{H}}(n, m) \leq \left( \frac{emn}{d + 1} \right)^{l(d+1)} \left( \frac{emn}{l + 1} \right)^{k(l+1)} \left( \frac{emn}{n(k + 1)} \right)^{n(k+1)}.$$

$f(x) := x \log x$ is a convex function, hence for all $a, b, c > 0$,

$$f \left( \frac{ka + lb + c}{k + l + 1} \right) \leq \frac{1}{k + l + 1} \left( k f(a) + l f(b) + f(c) \right)$$

$$\Rightarrow \quad \left( \frac{k + l + 1}{ka + lb + c} \right)^{ka + lb + c} \geq \left( \frac{1}{a} \right)^{ka} \left( \frac{1}{b} \right)^{lb} \left( \frac{1}{c} \right)^{c}.$$

Substituting $a = l + 1$, $b = d + 1$ and $c = n(k + 1)$ shows that

$$\Pi_{\mathbb{H}}(n, m) \leq \left( \frac{emn(k + l + 1)}{W + n(k + 1)} \right)^{W + n(k+1)}. \tag{51}$$

Hence, if

$$m > \left( \frac{W}{n} + k + 1 \right) \log_2 \left( \frac{emn(k+l+1)}{W + n(k+1)} \right) \tag{52}$$

then $\Pi_{\mathbb{H}}(n, m) < 2^{nm}$ and so by definition $d_{\mathbb{H}}(n) \leq m$. For all $a > 1$, observe that $x > a \log_2 x$ if $x = 2a \log_2 2a$. Setting $x = emn(k+l+1)/(W + n(k+1))$ and $a = e(k+l+1)$ shows that (52) is satisfied if $m = 2(W/n + k + 1) \log_2(2e(k+l+1))$. ∎

**Theorem 16.** *Let $\mathbb{H}$ be as in Theorem 15 with the following extra restrictions: $d \geq 3$, $l \geq k$ and $k \leq d$. Then*

$$d_{\mathbb{H}}(n) \geq \frac{1}{2} \left( \left\lfloor \frac{W}{2n} \right\rfloor + k + 1 \right)$$

*Proof.* We bound $\overline{d}(\mathbb{H})$ and $\underline{d}(\mathbb{H})$ and then apply Lemma 10. In the present setting $\mathbb{H}^1$ contains all three-layer linear-threshold networks with $d$ input nodes, $l$ hidden nodes in the first hidden layer, $k$ hidden nodes in the second hidden layer and one output node. From Theorem 13 in Bartlett (1993), we have

$$\mathrm{VCdim}(\mathbb{H}^1) \geq dl + \frac{l(k-1)}{2} + 1,$$

which under the restrictions stated above is greater than $W/2$. Hence $\overline{d}(\mathbb{H}) \geq W/2$.

As $k \leq d$ and $l \geq k$ we can choose a feature weight assignment so that the feature map is the identity on $k$ components of the input vector and insensitive to the setting of the remaining $d - k$ components. Hence we can generate $k + 1$ points in $X$ whose image under the feature map is shattered by the linear threshold output node, and so $\underline{d}(\mathbb{H}) = k + 1$. ∎

Combining Theorem 15 with Corrolary 13 shows that

$$m \geq O\left( \frac{1}{\varepsilon^2} \left[ \left( \frac{W}{n} + k + 1 \right) \log \frac{1}{\varepsilon} + \frac{1}{n} \log \frac{1}{\delta} \right] \right)$$

examples of each task suffice when learning $n$ tasks using a linear threshold hypothesis space family, while combining Theorem 16 with Theorem 14 shows that if

$$m \leq \Omega\left( \frac{1}{\varepsilon^2} \left[ \left( \frac{W}{n} + k + 1 \right) + \frac{1}{n} \log \frac{1}{\delta} \right] \right)$$

then any learning algorithm will fail on some set of $n$ tasks.

## 4. Conclusion

The problem of inductive bias is one that has broad significance in machine learning. In this paper we have introduced a formal model of inductive bias learning that applies when the learner is able to sample from multiple related tasks. We proved that provided certain covering numbers computed from the set of all hypothesis spaces available to the bias learner are finite, any hypothesis space that contains good solutions to sufficiently many training tasks is likely to contain good solutions to novel tasks drawn from the same environment.

In the specific case of learning a set of features, we showed that the number of examples $m$ required of each task in an $n$-task training set obeys $m = O(k + W/n)$, where $k$ is the number of features and $W$ is a measure of the complexity of the feature class. We showed that this bound is essentially tight for Boolean feature maps constructed from linear threshold networks. In addition, we proved that the number of tasks required to ensure good performance from the features on novel tasks is no more than $O(W)$. We also showed how a good set of features may be found by gradient descent.

The model of this paper represents a first step towards a formal model of hierarchical approaches to learning. By modelling a learner's uncertainty concerning its environment in probabilistic terms, we have shown how learning can occur simultaneously at both the base level—learn the tasks at hand—and at the meta-level—learn bias that can be transferred to novel tasks. From a technical perspective, it is the assumption that tasks are distributed probabilstically that allows the performance guarantees to be proved. From a practical perspective, there are many problem domains that can be viewed as probabilistically distributed sets of related tasks. For example, speech recognition may be decomposed along many different axes: words, speakers, accents, etc. Face recognition represents a potentially infinite domain of related tasks. Medical diagnosis and prognosis problems using the same pathology tests are yet another example. All of these domains should benefit from being tackled with a bias learning approach.

Natural avenues for further enquiry include:

- **Alternative constructions for** $\mathbb{H}$. Although widely applicable, the specific example on feature learning via gradient descent represents just one possible way of generating and searching the hypothesis space family $\mathbb{H}$. It would be interesting to investigate alternative methods, including decision tree approaches, approaches from Inductive Logic Programming (Khan et al., 1998), and whether more general learning techniques such as boosting can be applied in a bias learning setting.

- **Algorithms for automatically determining the hypothesis space family** $\mathbb{H}$. In our model the structure of $\mathbb{H}$ is fixed *apriori* and represents the *hyper-bias* of the bias learner. It would be interesting to see to what extent this structure can also be learnt.

- **Algorithms for automatically determining task relatedness.** In ordinary learning there is usually little doubt whether an individual *example* belongs to the same learning task or not. The analogous question in bias learning is whether an individual learning task belongs to a given set of related tasks, which in contrast to ordinary learning, does not always have such a clear-cut answer. For most of the examples we have discussed here, such as speech and face recognition, the task-relatedness is not in question, but in other cases such as medical problems it is not so clear. Grouping too large a subset of tasks together as related tasks could clearly have a detrimental impact on bias-learning or multi-task learning, and there is empirical evidence to support this (Caruana, 1997). Thus, algorithms for automatically determining task-relatedness are a potentially useful avenue for further research. In this context, see Silver and Mercer (1996), Thrun and O'Sullivan (1996). Note that the question of task relatedness is clearly only meaningful *relative* to a particular hypothesis space family $\mathbb{H}$ (for example, all possible collections of tasks are related if $\mathbb{H}$ contains every possible hypothesis space).

- **Extended hierarchies.** For an extension of our two-level approach to arbitrarily deep hierarchies, see Langford (1999). An interesting further question is to what extent the hierarchy can be inferred from data. This is somewhat related to the question of automatic induction of structure in graphical models.

## Acknowledgements

This work was supported at various times by an Australian Postgraduate Award, a Shell Australia Postgraduate Fellowship, U.K Engineering and Physical Sciences Research Council grants K70366 and K70373, and an Australian Postdoctoral Fellowship. Along the way, many people have contributed helpful comments and suggestions for improvement including Martin Anthony, Peter Bartlett, Rich Caruana, John Langford, Stuart Russell, John Shawe-Taylor, Sebastian Thrun and several anonymous referees.

## Appendix A. Uniform Convergence Results

Theorem 2 provides a bound (uniform over all $\mathcal{H} \in \mathbb{H}$) on the probability of large deviation between $\mathrm{er}_Q(\mathcal{H})$ and $\hat{\mathrm{er}}_{\mathbf{z}}(\mathcal{H})$. To obtain a more general result, we follow Haussler (1992) and introduce the following parameterized class of metrics on $\mathbb{R}^+$:

$$d_\nu[x, y] := \frac{|x - y|}{x + y + \nu},$$

where $\nu > 0$. Our main theorem will be a uniform bound on the probability of large values of $d_\nu[\mathrm{er}_Q(\mathcal{H}), \mathrm{er}_{\mathbf{z}}(\mathcal{H})]$, rather than $|\mathrm{er}_Q(\mathcal{H}) - \hat{\mathrm{er}}_{\mathbf{z}}(\mathcal{H})|$. Theorem 2 will then follow as a corollary, as will better bounds for the realizable case $\hat{\mathrm{er}}_{\mathbf{z}}(\mathcal{H}) = 0$ (Appendix A.3).

**Lemma 17.** *The following three properties of $d_\nu$ are easily established:*

1. *For all $r, s \geq 0$, $0 \leq d_\nu[r, s] \leq 1$*

2. *For all $0 \leq r \leq s \leq t$, $d_\nu[r, s] \leq d_\nu[r, t]$ and $d_\nu[s, t] \leq d_\nu[r, t]$.*

3. *For $0 \leq r, s \leq 1$, $\frac{|r-s|}{\nu+2} \leq d_\nu[r, s] \leq \frac{|r-s|}{\nu}$*

For ease of exposition we have up until now been dealing explicitly with hypothesis spaces $\mathcal{H}$ containing functions $h \colon X \to Y$, and then constructing loss functions $h_l$ mapping $X \times Y \to [0, 1]$ by $h_l(x, y) := l(h(x), y)$ for some loss function $l \colon Y \times Y \to [0, 1]$. However, in general we can view $h_l$ just as a function from an abstract set $Z$ $(X \times Y)$ to $[0, 1]$ and ignore its particular construction in terms of the loss function $l$. So for the remainder of this section, unless otherwise stated, all hypothesis spaces $\mathcal{H}$ will be sets of functions mapping $Z$ to $[0, 1]$. It will also be considerably more convenient to transpose our notation for $(n, m)$-samples, writing the $n$ training sets as columns instead of rows:

$$\mathbf{z} = \begin{matrix} z_{11} & \ldots & z_{1n} \\ \vdots & \ddots & \vdots \\ z_{m1} & \ldots & z_{mn} \end{matrix}$$

where each $z_{ij} \in Z$. Recalling the definition of $(X \times Y)^{(n,m)}$ (Equation 9 and prior discussion), with this transposition $\mathbf{z}$ lives in $(X \times Y)^{(m,n)}$. The following definition now generalizes quantities like $\hat{\mathrm{er}}_{\mathbf{z}}(\mathcal{H})$, $\mathrm{er}_{\mathbf{P}}(\mathcal{H})$ and so on to this new setting.

**Definition 6.** Let $\mathcal{H}_1, \ldots, \mathcal{H}_n$ be $n$ sets of functions mapping $Z$ into $[0, 1]$. For any $h_1 \in \mathcal{H}_1, \ldots, h_n \in \mathcal{H}_n$, let $h_1 \oplus \cdots \oplus h_n$ or simply $\mathbf{h}$ denote the map

$$\mathbf{h}(\vec{z}) = 1/n \sum_{i=1}^n h_i(z_i)$$

for all $\vec{z} = (z_1, \ldots, z_n) \in Z^n$. Let $\mathcal{H}_1 \oplus \cdots \oplus \mathcal{H}_n$ denote the set of all such functions. Given $\mathbf{h} \in \mathcal{H}_1 \oplus \cdots \oplus \mathcal{H}_n$ and $m$ elements of $(X \times Y)^n$, $(\vec{z}_1, \ldots, \vec{z}_m)$ (or equivalently an element $\mathbf{z}$ of $(X \times Y)^{(m,n)}$ by writing the $\vec{z}_i$ as rows), define

$$\hat{\mathrm{er}}_{\mathbf{Z}}(\mathbf{h}) := \frac{1}{m} \sum_{i=1}^{m} \mathbf{h}(\vec{z}_i)$$

(recall equation (8)). Similarly, for any product probability measure $\mathbf{P} = P_1 \times \cdots \times P_n$ on $(X \times Y)^n$, define

$$\mathrm{er}_{\mathbf{P}}(\mathbf{h}) := \int_{Z^n} \mathbf{h}(\vec{z}) \, d\mathbf{P}(\vec{z})$$

(recall equation (26)). For *any* $\mathbf{h}, \mathbf{h}' \colon (X \times Y)^n \to [0, 1]$ (not necessarily of the form $h_1 \oplus \cdots \oplus h_n$), define

$$d_{\mathbf{P}}(\mathbf{h}, \mathbf{h}') := \int_{Z^n} |\mathbf{h}(\vec{z}) - \mathbf{h}'(\vec{z})| \, d\mathbf{P}(\vec{z})$$

(recall equation (17)). For any class of functions $\mathcal{H}$ mapping $(X \times Y)^n$ to $[0, 1]$, define

$$\mathcal{C}(\varepsilon, \mathcal{H}) := \sup_{\mathbf{P}} \mathcal{N}(\varepsilon, \mathcal{H}, d_{\mathbf{P}})$$

where the supremum is over all product probability measures on $(X \times Y)^n$ and $\mathcal{N}(\varepsilon, \mathcal{H}, d_{\mathbf{P}})$ is the size of the smallest $\varepsilon$-cover of $\mathcal{H}$ under $d_{\mathbf{P}}$ (recall Definition 4).

The following theorem is the main result from which the rest of the uniform convergence results in this paper are derived.

**Theorem 18.** *Let $\mathcal{H} \subseteq \mathcal{H}_1 \oplus \cdots \oplus \mathcal{H}_n$ be a permissible class of functions mapping $(X \times Y)^n$ into $[0, 1]$. Let $\mathbf{z} \in (X \times Y)^{(m,n)}$ be generated by $m \geq 2/(\alpha^2 \nu)$ independent trials from $(X \times Y)^n$ according to some product probability measure $\mathbf{P} = P_1 \times \cdots \times P_n$. For all $\nu > 0$, $0 < \alpha < 1$,*

$$\Pr\left\{\mathbf{z} \in (X \times Y)^{(m,n)} \colon \sup_{\mathcal{H}} d_\nu\left[\hat{\mathrm{er}}_{\mathbf{Z}}(\mathbf{h}), \mathrm{er}_{\mathbf{P}}(\mathbf{h})\right] > \alpha\right\}$$
$$\leq 4\mathcal{C}(\alpha\nu/8, \mathcal{H}) \exp(-\alpha^2 \nu n m / 8). \quad (53)$$

The following immediate corollary will also be of use later.

**Corollary 19.** *Under the same conditions as Theorem 18, if*

$$m \geq \max\left\{\frac{8}{\alpha^2 \nu n} \log \frac{4\mathcal{C}\left(\frac{\alpha\nu}{8}, \mathcal{H}\right)}{\delta}, \frac{2}{\alpha^2 \nu}\right\}, \quad (54)$$

*then*

$$\Pr\left\{\mathbf{z} \in (X \times Y)^{(m,n)} \colon \sup_{\mathcal{H}} d_\nu\left[\hat{\mathrm{er}}_{\mathbf{Z}}(\mathbf{h}), \mathrm{er}_{\mathbf{P}}(\mathbf{h})\right] > \alpha\right\} \leq \delta \quad (55)$$

### A.1 Proof of Theorem 18

The proof is via a double symmetrization argument of the kind given in chapter 2 of Pollard (1984). I have also borrowed some ideas from the proof of Theorem 3 in Haussler (1992).

### A.1.1 FIRST SYMMETRIZATION

An extra piece of notation: for all $\mathbf{z} \in (X \times Y)^{(2m,n)}$, let $\mathbf{z}(1)$ be the top half of $\mathbf{z}$ and $\mathbf{z}(2)$ be the bottom half, viz:

$$\mathbf{z}(1) = \begin{matrix} z_{11} & \dots & z_{1n} \\ \vdots & \ddots & \vdots \\ z_{m1} & \dots & z_{mn} \end{matrix} \qquad \mathbf{z}(2) = \begin{matrix} z_{m+1,1} & \dots & z_{m+1,n} \\ \vdots & \ddots & \vdots \\ z_{2m,1} & \dots & z_{2m,n} \end{matrix}$$

The following lemma is the first "symmetrization trick." We relate the probability of large deviation between an empirical estimate of the loss and the true loss to the probability of large deviation between two independent empirical estimates of the loss.

**Lemma 20.** *Let $\mathcal{H}$ be a permissible set of functions from $(X \times Y)^n$ into $[0,1]$ and let $P$ be a probability measure on $(X \times Y)^n$. For all $\nu > 0, 0 < \alpha < 1$ and $m \geq \frac{2}{\alpha^2 \nu}$,*

$$\Pr\left\{\mathbf{z} \in (X \times Y)^{(m,n)} \colon \sup_{\mathcal{H}} d_\nu\left[\hat{\mathrm{er}}_{\mathbf{z}}(h), \mathrm{er}_P(h)\right] > \alpha\right\}$$
$$\leq 2 \Pr\left\{\mathbf{z} \in Z^{(2m,n)} \colon \sup_{\mathcal{H}} d_\nu\left[\hat{\mathrm{er}}_{\mathbf{z}(1)}(h), \hat{\mathrm{er}}_{\mathbf{z}(2)}(h)\right] > \frac{\alpha}{2}\right\}. \quad (56)$$

*Proof.* Note first that permissibility of $\mathcal{H}$ guarantees the measurability of suprema over $\mathcal{H}$ (Lemma 32 part 5). By the triangle inequality for $d_\nu$, if $d_\nu\left[\hat{\mathrm{er}}_{\mathbf{z}(1)}(h), \mathrm{er}_P(h)\right] > \alpha$ and $d_\nu\left[\hat{\mathrm{er}}_{\mathbf{z}(2)}(h), \mathrm{er}_P(h)\right] < \alpha/2$, then $d_\nu\left[\hat{\mathrm{er}}_{\mathbf{z}(1)}(h), \hat{\mathrm{er}}_{\mathbf{z}(2)}(h)\right] > \alpha/2$. Thus,

$$\Pr\left\{\mathbf{z} \in (X \times Y)^{(2m,n)} \colon \exists h \in \mathcal{H} \colon d_\nu\left[\hat{\mathrm{er}}_{\mathbf{z}(1)}(h), \hat{\mathrm{er}}_{\mathbf{z}(2)}(h)\right] > \frac{\alpha}{2}\right\}$$
$$\geq \Pr\Big\{\mathbf{z} \in (X \times Y)^{(2m,n)} \colon \exists h \in \mathcal{H} \colon d_\nu\left[\hat{\mathrm{er}}_{\mathbf{z}(1)}(h), \mathrm{er}_P(h)\right] > \alpha \text{ and }$$
$$d_\nu\left[\hat{\mathrm{er}}_{\mathbf{z}(2)}(h), \mathrm{er}_P(h)\right] < \alpha/2\Big\}. \quad (57)$$

By Chebyshev's inequality, for any fixed $h$,

$$\Pr\left\{\mathbf{z} \in (X \times Y)^{(m,n)} \colon d_\nu\left[\hat{\mathrm{er}}_{\mathbf{z}}(h), \mathrm{er}_P(h)\right] < \frac{\alpha}{2}\right\}$$
$$\geq \Pr\left\{\mathbf{z} \in (X \times Y)^{(m,n)} \colon \frac{|\hat{\mathrm{er}}_{\mathbf{z}}(h) - \mathrm{er}_P(h)|}{\nu} < \frac{\alpha}{2}\right\}$$
$$\geq 1 - \frac{\mathrm{er}_P(h)(1 - \mathrm{er}_P(h))}{m\nu\alpha^2/4}$$
$$\geq \frac{1}{2}$$

as $m \geq 2/(\alpha^2 \nu)$ and $\mathrm{er}_P(h) \leq 1$. Substituting this last expression into the right hand side of (57) gives the result. $\qquad \square$

### A.1.2 SECOND SYMMETRIZATION

The second symmetrization trick bounds the probability of large deviation between two empirical estimates of the loss (i.e. the right hand side of (56)) by computing the probability of large deviation when elements are randomly permuted between the first and second sample. The following definition introduces the appropriate permutation group for this purpose.

**Definition 7.** For all integers $m, n \geq 1$, let $\Gamma_{(2m,n)}$ denote the set of all permutations $\sigma$ of the sequence of pairs of integers $\{(1,1), \ldots, (1, n), \ldots, (2m, 1), \ldots, (2m, n)\}$ such that for all $i$, $1 \leq i \leq m$, either $\sigma(i, j) = (m + i, j)$ and $\sigma(m + i, j) = (i, j)$ or $\sigma(i, j) = (i, j)$ and $\sigma(m + i, j) = (m + i, j)$.

For any $\mathbf{z} \in (X \times Y)^{(2m,n)}$ and any $\sigma \in \Gamma_{(2m,n)}$, let

$$\mathbf{z}_\sigma := \begin{matrix} z_{\sigma(1,1)} & \cdots & z_{\sigma(1,n)} \\ \vdots & \ddots & \vdots \\ z_{\sigma(2m,1)} & \cdots & z_{\sigma(2m,n)}. \end{matrix}$$

**Lemma 21.** Let $\mathcal{H} = \mathcal{H}_1 \oplus \cdots \oplus \mathcal{H}_n$ be a permissible set of functions mapping $(X \times Y)^n$ into $[0, 1]$ (as in the statement of Theorem 18). Fix $\mathbf{z} \in (X \times Y)^{(2m,n)}$ and let $\hat{\mathcal{H}} := \{\mathbf{f}^1, \ldots, \mathbf{f}^M\}$ be an $\alpha\nu/8$-cover for $(\mathcal{H}, d_\mathbf{z})$, where $d_\mathbf{z}(\mathbf{h}, \mathbf{h}') := \frac{1}{2m} \sum_{i=1}^{2m} |\mathbf{h}(\vec{z}_i) - \mathbf{h}'(\vec{z}_i)|$ where the $\vec{z}_i$ are the rows of $\mathbf{z}$. Then,

$$\Pr\left\{ \sigma \in \Gamma_{(2m,n)} \colon \sup_{\mathcal{H}} d_\nu\left[\hat{er}_{\mathbf{z}_{\sigma(1)}}(\mathbf{h}), \hat{er}_{\mathbf{z}_{\sigma(2)}}(\mathbf{h})\right] > \frac{\alpha}{2} \right\}$$

$$\leq \sum_{i=1}^{M} \Pr\left\{ \sigma \in \Gamma_{(2m,n)} \colon d_\nu\left[\hat{er}_{\mathbf{z}_{\sigma(1)}}(\mathbf{f}^i), \hat{er}_{\mathbf{z}_{\sigma(2)}}(\mathbf{f}^i)\right] > \frac{\alpha}{4} \right\}, \quad (58)$$

where each $\sigma \in \Gamma_{(2m,n)}$ is chosen uniformly at random.

*Proof.* Fix $\sigma \in \Gamma_{(2m,n)}$ and let $\mathbf{h} \in \mathcal{H}$ be such that $d_\nu\left[\hat{er}_{\mathbf{z}_{\sigma(1)}}(\mathbf{h}), \hat{er}_{\mathbf{z}_{\sigma(2)}}(\mathbf{h})\right] > \alpha/2$ (if there is no such $\mathbf{h}$ for any $\sigma$ we are already done). Choose $\mathbf{f} \in \hat{\mathcal{H}}$ such that $d_\mathbf{z}(\mathbf{h}, \mathbf{f}) \leq \alpha\nu/8$. Without loss of generality we can assume $\mathbf{f}$ is of the form $\mathbf{f} = f_1 \oplus \cdots \oplus f_n$. Now,

$$\frac{2}{\nu} d_\mathbf{z}(\mathbf{h}, \mathbf{f}) = \frac{\sum_{i=1}^{2m} \left| \sum_{j=1}^{n} h_j(z_{ij}) - f_j(z_{ij}) \right|}{\nu m n}$$

$$= \frac{\sum_{i=1}^{2m} \left| \sum_{j=1}^{n} h_j(z_{\sigma(i,j)}) - f_j(z_{\sigma(i,j)}) \right|}{\nu m n}$$

$$\geq \frac{\left| \sum_{i=1}^{m} \sum_{j=1}^{n} \left[ h_j(z_{\sigma(i,j)}) - f_j(z_{\sigma(i,j)}) \right] \right|}{\nu m n + \sum_{i=1}^{m} \sum_{j=1}^{n} \left[ h_j(z_{\sigma(i,j)}) + f_j(z_{\sigma(i,j)}) \right]}$$

$$\qquad + \frac{\left| \sum_{i=m+1}^{2m} \sum_{j=1}^{n} \left[ h_j(z_{\sigma(i,j)}) - f_j(z_{\sigma(i,j)}) \right] \right|}{\nu m n + \sum_{i=m+1}^{2m} \sum_{j=1}^{n} \left[ h_j(z_{\sigma(i,j)}) + f_j(z_{\sigma(i,j)}) \right]}$$

$$= d_\nu\left[\hat{er}_{\mathbf{z}_{\sigma(1)}}(\mathbf{h}), \hat{er}_{\mathbf{z}_{\sigma(1)}}(\mathbf{f})\right] + d_\nu\left[\hat{er}_{\mathbf{z}_{\sigma(2)}}(\mathbf{h}), \hat{er}_{\mathbf{z}_{\sigma(2)}}(\mathbf{f})\right].$$

Hence, by the triangle inequality for $d_\nu$,

$$
\begin{aligned}
\frac{2}{\nu} d_{\mathbf{z}}(\mathbf{h}, \mathbf{f}) + d_\nu \left[ \dot{\mathrm{er}}_{\mathbf{z}_{\sigma(1)}}(\mathbf{f}), \dot{\mathrm{er}}_{\mathbf{z}_{\sigma(2)}}(\mathbf{f}) \right] &\geq d_\nu \left[ \dot{\mathrm{er}}_{\mathbf{z}_{\sigma(1)}}(\mathbf{h}), \dot{\mathrm{er}}_{\mathbf{z}_{\sigma(1)}}(\mathbf{f}) \right] \\
&+ d_\nu \left[ \dot{\mathrm{er}}_{\mathbf{z}_{\sigma(2)}}(\mathbf{h}), \dot{\mathrm{er}}_{\mathbf{z}_{\sigma(2)}}(\mathbf{f}) \right] + d_\nu \left[ \dot{\mathrm{er}}_{\mathbf{z}_{\sigma(1)}}(\mathbf{f}), \dot{\mathrm{er}}_{\mathbf{z}_{\sigma(2)}}(\mathbf{f}) \right] \\
&\geq d_\nu \left[ \dot{\mathrm{er}}_{\mathbf{z}_{\sigma(1)}}(\mathbf{h}), \dot{\mathrm{er}}_{\mathbf{z}_{\sigma(2)}}(\mathbf{h}) \right] .
\end{aligned}
\tag{59}
$$

But $\frac{2}{\nu} d_{\mathbf{z}}(\mathbf{h}, \mathbf{f}) \leq \alpha/4$ by construction and $d_\nu \left[ \dot{\mathrm{er}}_{\mathbf{z}_{\sigma(1)}}(\mathbf{h}), \dot{\mathrm{er}}_{\mathbf{z}_{\sigma(2)}}(\mathbf{h}) \right] > \alpha/2$ by assumption, so (59) implies $d_\nu \left[ \dot{\mathrm{er}}_{\mathbf{z}_{\sigma(1)}}(\mathbf{f}), \dot{\mathrm{er}}_{\mathbf{z}_{\sigma(2)}}(\mathbf{f}) \right] > \alpha/4$. Thus,

$$
\begin{aligned}
\left\{ \sigma \in \Gamma_{(2m,n)} \colon \exists \mathbf{h} \in \mathcal{H} \colon d_\nu \left[ \dot{\mathrm{er}}_{\mathbf{z}_{\sigma(1)}}(\mathbf{h}), \dot{\mathrm{er}}_{\mathbf{z}_{\sigma(2)}}(\mathbf{h}) \right] > \frac{\alpha}{2} \right\} \\
\subseteq \left\{ \sigma \in \Gamma_{(2m,n)} \colon \exists \mathbf{f} \in \hat{\mathcal{H}} \colon d_\nu \left[ \dot{\mathrm{er}}_{\mathbf{z}_{\sigma(1)}}(\mathbf{f}), \dot{\mathrm{er}}_{\mathbf{z}_{\sigma(2)}}(\mathbf{f}) \right] \right) > \frac{\alpha}{4} \right\},
\end{aligned}
$$

which gives (58). □

Now we bound the probability of each term in the right hand side of (58).

**Lemma 22.** *Let* $\mathbf{f} \colon (X \times Y)^n \to [0,1]$ *be any function that can be written in the form* $\mathbf{f} = f_1 \oplus \cdots \oplus f_n$. *For any* $\mathbf{z} \in (X \times Y)^{(2m,n)}$,

$$
\Pr \left\{ \sigma \in \Gamma_{(2m,n)} \colon d_\nu \left[ \dot{\mathrm{er}}_{\mathbf{z}_{\sigma(1)}}(\mathbf{f}), \dot{\mathrm{er}}_{\mathbf{z}_{\sigma(2)}}(\mathbf{f}) \right] > \frac{\alpha}{4} \right\} \leq 2 \exp \left( \frac{-\alpha^2 \nu m n}{8} \right),
\tag{60}
$$

*where each* $\sigma \in \Gamma_{(2m,n)}$ *is chosen uniformly at random.*

*Proof.* For any $\sigma \in \Gamma_{(2m,n)}$,

$$
d_\nu \left[ \dot{\mathrm{er}}_{\mathbf{z}_{\sigma(1)}}(\mathbf{f}), \dot{\mathrm{er}}_{\mathbf{z}_{\sigma(2)}}(\mathbf{f}) \right] = \frac{\left| \sum_{i=1}^m \sum_{j=1}^n \left[ f_j(z_{\sigma(i,j)}) - f_j(z_{\sigma(m+i,j)}) \right] \right|}{\nu m n + \sum_{i=1}^{2m} \sum_{j=1}^n f_j(z_{ij})}.
\tag{61}
$$

To simplify the notation denote $f_j(z_{ij})$ by $\beta_{ij}$. For each pair $ij$, $1 \leq i \leq m$, $1 \leq j \leq n$, let $Y_{ij}$ be an independent random variable such that $Y_{ij} = \beta_{ij} - \beta_{m+i,j}$ with probability $1/2$ and $Y_{ij} = \beta_{m+i,j} - \beta_{ij}$ with probability $1/2$. From (61),

$$
\begin{aligned}
&\Pr \left\{ \sigma \in \Gamma_{(2m,n)} \colon d_\nu \left[ \dot{\mathrm{er}}_{\mathbf{z}_{\sigma(1)}}(\mathbf{f}), \dot{\mathrm{er}}_{\mathbf{z}_{\sigma(2)}}(\mathbf{f}) \right] > \frac{\alpha}{4} \right\} \\
&= \Pr \left\{ \sigma \in \Gamma_{(2m,n)} \colon \left| \sum_{i=1}^m \sum_{j=1}^n \left[ f_j(z_{\sigma(i,j)}) - f_j(z_{\sigma(m+i,j)}) \right] \right| > \frac{\alpha}{4} \left( \nu m n + \sum_{i=1}^{2m} \sum_{j=1}^n \beta_{ij} \right) \right\} \\
&= \Pr \left\{ \left| \sum_{i=1}^m \sum_{j=1}^n Y_{ij} \right| \frac{\alpha}{4} \left( \nu m n + \sum_{i=1}^{2m} \sum_{j=1}^n \beta_{ij} \right) \right\}
\end{aligned}
$$

For zero-mean independent random variables $Y_1, \ldots, Y_k$ with bounded ranges $a_i \leq Y_i \leq b_i$, Hoeffding's inequality (Devroye, Györfi, & Lugosi, 1996) is

$$
\Pr \left\{ \left| \sum_{i=1}^k Y_i \right| \geq \eta \right\} \leq 2 \exp \left( -\frac{2\eta^2}{\sum_{i=1}^k (b_i - a_i)^2} \right).
$$

Noting that the range of each $Y_{ij}$ is $[-|\beta_{ij} - \beta_{i+m,j})|, |\beta_{ij} - \beta_{i+m,j})|]$, we have

$$\Pr\left\{\left|\sum_{i=1}^{m}\sum_{j=1}^{n} Y_{ij}\right| > \frac{\alpha}{4}\left(\nu mn + \sum_{i=1}^{2m}\sum_{j=1}^{n}\beta_{ij}\right)\right\} \le 2\exp\left(-\frac{\alpha^2\left[\nu mn + \sum_{i=1}^{2m}\sum_{j=1}^{n}\beta_{ij}\right]^2}{32\sum_{i=1}^{m}\sum_{j=1}^{n}(\beta_{ij} - \beta_{i+m,j})^2}\right)$$

Let $\gamma = \sum_{i=1}^{2m}\sum_{j=1}^{n}\beta_{ij}$. As $0 \le \beta_{ij} \le 1$, $\sum_{i=1}^{m}\sum_{j=1}^{n}(\beta_{ij} - \beta_{m+ij})^2 \le \gamma$. Hence,

$$2\exp\left(-\frac{\alpha^2\left[\nu mn + \sum_{i=1}^{2m}\sum_{j=1}^{n}\beta_{ij}\right]^2}{32\sum_{i=1}^{m}\sum_{j=1}^{n}(\beta_{ij} - \beta_{i+m,j})^2}\right) \le 2\exp\left(-\frac{\alpha^2(\nu mn + \gamma)^2}{32\gamma}\right).$$

$(\nu mn + \gamma)^2/\gamma$ is minimized by setting $\gamma = \nu mn$ giving a value of $4\nu mn$. Hence

$$\Pr\left\{\sigma \in \Gamma_{(2m,n)}\colon d_\nu\left[\hat{\mathrm{er}}_{\mathbf{z}_{\sigma(1)}}(\mathbf{f}), \hat{\mathrm{er}}_{\mathbf{z}_{\sigma(2)}}(\mathbf{f})\right] > \frac{\alpha}{4}\right\} \le 2\exp\left(\frac{-\alpha^2\nu mn}{8}\right),$$

as required. $\qquad\square$

### A.1.3 PUTTING IT TOGETHER

For fixed $\mathbf{z} \in (X \times Y)^{(2m,n)}$, Lemmas 21 and 22 give:

$$\Pr\left\{\sigma \in \Gamma_{(2m,n)}\colon \sup_{\mathcal{H}} d_\nu\left[\hat{\mathrm{er}}_{\mathbf{z}_{\sigma(1)}}(\mathbf{h}), \hat{\mathrm{er}}_{\mathbf{z}_{\sigma(2)}}(\mathbf{h})\right] > \frac{\alpha}{2}\right\}$$
$$\le 2\mathcal{N}\left(\alpha\nu/8, \mathcal{H}, d_{\mathbf{z}}\right)\exp\left(-\frac{\alpha^2\nu mn}{8}\right).$$

Note that $d_{\mathbf{z}}$ is simply $d_{\mathbf{P}}$ where $\mathbf{P} = (P_1, \ldots, P_n)$ and each $P_i$ is the empirical distribution that puts point mass $1/m$ on each $z_{ji}, j = 1, \ldots, 2m$ (recall Definition 3). Hence,

$$\Pr\left\{\sigma \in \Gamma_{(2m,n)}, \mathbf{z} \in (X \times Y)^{(2m,n)}\colon \sup_{\mathcal{H}} d_\nu\left[\hat{\mathrm{er}}_{\mathbf{z}_{\sigma(1)}}(\mathbf{h}), \hat{\mathrm{er}}_{\mathbf{z}_{\sigma(2)}}(\mathbf{h})\right] > \frac{\alpha}{2}\right\}$$
$$\le 2\mathcal{C}\left(\alpha\nu/8, \mathcal{H}\right)\exp\left(-\frac{\alpha^2\nu mn}{8}\right).$$

Now, for a random choice of $\mathbf{z}$, each $z_{ij}$ in $\mathbf{z}$ is independently (but not identically) distributed and $\sigma$ only ever swaps $z_{ij}$ and $z_{i+m,j}$ (so that $\sigma$ swaps a $z_{ij}$ drawn according to $P_j$ with another component drawn according to the same distribution). Thus we can integrate out with respect to the choice of $\sigma$ and write

$$\Pr\left\{\mathbf{z} \in (X \times Y)^{(2m,n)}\colon \sup_{\mathcal{H}} d_\nu\left[\hat{\mathrm{er}}_{\mathbf{z}(1)}(\mathbf{h}), \hat{\mathrm{er}}_{\mathbf{z}(2)}(\mathbf{h})\right] > \frac{\alpha}{2}\right\}$$
$$\le 2\mathcal{C}\left(\alpha\nu/8, \mathcal{H}\right)\exp\left(-\frac{\alpha^2\nu mn}{8}\right).$$

Applying Lemma 20 to this expression gives Theorem 18. $\qquad\square$

### A.2 Proof of Theorem 2

Another piece of notation is required for the proof. For any hypothesis space $\mathcal{H}$ and any probability measures $\mathbf{P} = (P_1, \ldots, P_n)$ on $Z$, let

$$\acute{\mathrm{er}}_{\mathbf{P}}(\mathcal{H}) := \frac{1}{n} \sum_{i=1}^{n} \inf_{h \in \mathcal{H}} \mathrm{er}_{P_i}(h).$$

Note that we have used $\acute{\mathrm{er}}_{\mathbf{P}}(\mathcal{H})$ rather than $\mathrm{er}_{\mathbf{P}}(\mathcal{H})$ to indicate that $\acute{\mathrm{er}}_{\mathbf{P}}(\mathcal{H})$ is another empirical estimate of $\mathrm{er}_Q(\mathcal{H})$.

With the $(n, m)$-sampling process, in addition to the sample $\mathbf{z}$ there is also generated a sequence of probability measures, $\mathbf{P} = (P_1, \ldots, P_n)$ although these are not supplied to the learner. This notion is used in the following Lemma, where $\Pr\{(\mathbf{z}, \mathbf{P}) \in (X \times Y)^{(n,m)} \times \mathcal{P}^n : A\}$ means "the probability of generating a sequence of measures $\mathbf{P}$ from the environment $(\mathcal{P}, Q)$ and then an $(n, m)$-sample $\mathbf{z}$ according to $\mathbf{P}$ such that A holds".

**Lemma 23.** *If*

$$\Pr\left\{(\mathbf{z}, \mathbf{P}) \in (X \times Y)^{(n,m)} \times \mathcal{P}^n \colon \sup_{\mathbb{H}} d_\nu \left[\acute{\mathrm{er}}_{\mathbf{z}}(\mathcal{H}), \acute{\mathrm{er}}_{\mathbf{P}}(\mathcal{H})\right] > \frac{\alpha}{2}\right\} \le \frac{\delta}{2}, \tag{62}$$

*and*

$$\Pr\left\{\mathbf{P} \in \mathcal{P}^n \colon \sup_{\mathbb{H}} d_\nu \left[\acute{\mathrm{er}}_{\mathbf{P}}(\mathcal{H}), \mathrm{er}_Q(\mathcal{H})\right] > \frac{\alpha}{2}\right\} \le \frac{\delta}{2}, \tag{63}$$

*then*

$$\Pr\left\{\mathbf{z} \in (X \times Y)^{(n,m)} \colon \sup_{\mathbb{H}} d_\nu \left[\acute{\mathrm{er}}_{\mathbf{z}}(\mathcal{H}), \mathrm{er}_Q(\mathcal{H})\right] > \alpha\right\} \le \delta.$$

*Proof.* Follows directly from the triangle inequality for $d_\nu$. ∎

We treat the two inequalities in Lemma 23 separately.

### A.2.1 Inequality (62)

In the following Lemma we replace the supremum over $\mathcal{H} \in \mathbb{H}$ in inequality (62) with a supremum over $\mathbf{h} \in \mathbb{H}^n$.

**Lemma 24.**

$$\Pr\left\{(\mathbf{z}, \mathbf{P}) \in (X \times Y)^{(n,m)} \times \mathcal{P}^n \colon \sup_{\mathbb{H}} d_\nu \left[\acute{\mathrm{er}}_{\mathbf{z}}(\mathcal{H}), \acute{\mathrm{er}}_{\mathbf{P}}(\mathcal{H})\right] > \alpha\right\}$$

$$\le \Pr\left\{(\mathbf{z}, \mathbf{P}) \in (X \times Y)^{(n,m)} \times \mathcal{P}^n \colon \sup_{\mathbb{H}_l^n} d_\nu \left[\acute{\mathrm{er}}_{\mathbf{z}}(\mathbf{h}), \mathrm{er}_{\mathbf{P}}(\mathbf{h})\right] > \alpha\right\} \tag{64}$$

*Proof.* Suppose that $(\mathbf{z}, \mathbf{P})$ are such that $\sup_{\mathbb{H}} d_\nu \left[\acute{\mathrm{er}}_{\mathbf{z}}(\mathcal{H}), \acute{\mathrm{er}}_{\mathbf{P}}(\mathcal{H})\right] > \alpha$. Let $\mathcal{H}$ satisfy this inequality. Suppose first that $\acute{\mathrm{er}}_{\mathbf{z}}(\mathcal{H}) \le \acute{\mathrm{er}}_{\mathbf{P}}(\mathcal{H})$. By the definition of $\acute{\mathrm{er}}_{\mathbf{z}}(\mathcal{H})$, for all $\varepsilon > 0$ there exists $\mathbf{h} \in \mathcal{H}^n := \mathcal{H} \oplus \cdots \oplus \mathcal{H}$ such that $\acute{\mathrm{er}}_{\mathbf{z}}(\mathbf{h}) < \acute{\mathrm{er}}_{\mathbf{z}}(\mathcal{H}) + \varepsilon$. Hence by property (3) of the $d_\nu$ metric, for all $\varepsilon > 0$, there exists $\mathbf{h} \in \mathcal{H}^n$ such that $d_\nu \left[\acute{\mathrm{er}}_{\mathbf{z}}(\mathbf{h}), \acute{\mathrm{er}}_{\mathbf{z}}(\mathcal{H})\right] < \varepsilon$. Pick an arbitrary $\mathbf{h}$ satisfying this inequality. By definition, $\acute{\mathrm{er}}_{\mathbf{P}}(\mathcal{H}) \le \mathrm{er}_{\mathbf{P}}(\mathbf{h})$, and so $\acute{\mathrm{er}}_{\mathbf{z}}(\mathcal{H}) \le \acute{\mathrm{er}}_{\mathbf{P}}(\mathcal{H}) \le \mathrm{er}_{\mathbf{P}}(\mathbf{h})$. As $d_\nu \left[\acute{\mathrm{er}}_{\mathbf{z}}(\mathcal{H}), \acute{\mathrm{er}}_{\mathbf{P}}(\mathcal{H})\right] > \alpha$ (by assumption), by the compatibility of $d_\nu$ with the ordering on the reals, $d_\nu \left[\acute{\mathrm{er}}_{\mathbf{z}}(\mathcal{H}), \mathrm{er}_{\mathbf{P}}(\mathbf{h})\right] > \alpha = \alpha + \delta$, say. By the triangle inequality for $d_\nu$,

$$d_\nu \left[\acute{\mathrm{er}}_{\mathbf{z}}(\mathbf{h}), \mathrm{er}_{\mathbf{P}}(\mathbf{h})\right] + d_\nu \left[\acute{\mathrm{er}}_{\mathbf{z}}(\mathbf{h}), \acute{\mathrm{er}}_{\mathbf{z}}(\mathcal{H})\right] \ge d_\nu \left[\acute{\mathrm{er}}_{\mathbf{z}}(\mathcal{H}), \mathrm{er}_{\mathbf{P}}(\mathbf{h})\right] = \alpha + \delta.$$

Thus $d_\nu \left[\acute{\mathrm{er}}_{\mathbf{z}}(\mathbf{h}), \mathrm{er}_{\mathbf{P}}(\mathbf{h})\right] > \alpha + \delta - \varepsilon$ and for any $\varepsilon > 0$ an $\mathbf{h}$ satisfying this inequality can be found. Choosing $\varepsilon = \delta$ shows that there exists $\mathbf{h} \in \mathcal{H}^n$ such that $d_\nu \left[\acute{\mathrm{er}}_{\mathbf{z}}(\mathbf{h}), \mathrm{er}_{\mathbf{P}}(\mathbf{h})\right] > \alpha$.

If instead, $\acute{\mathrm{er}}_{\mathbf{P}}(\mathcal{H}) < \acute{\mathrm{er}}_{\mathbf{z}}(\mathcal{H})$, then an identical argument can be run with the role of $\mathbf{z}$ and $\mathbf{P}$ interchanged. Thus in both cases,

$$\sup_{\mathbb{H}} d_\nu \left[\acute{\mathrm{er}}_{\mathbf{z}}(\mathcal{H}), \acute{\mathrm{er}}_{\mathbf{P}}(\mathcal{H})\right] > \alpha \Rightarrow \exists \mathbf{h} \in \mathbb{H}_l^n : d_\nu \left[\acute{\mathrm{er}}_{\mathbf{z}}(\mathbf{h}), \mathrm{er}_{\mathbf{P}}(\mathbf{h})\right] > \alpha,$$

which completes the proof of the Lemma. $\qquad\square$

By the nature of the $(n, m)$ sampling process,

$$\Pr \left\{ (\mathbf{z}, \mathbf{P}) \in (X \times Y)^{(n,m)} \times \mathcal{P}^n \sup_{\mathbb{H}_l^n} : d_\nu \left[\acute{\mathrm{er}}_{\mathbf{z}}(\mathbf{h}), \mathrm{er}_{\mathbf{P}}(\mathbf{h})\right] > \alpha \right\}$$

$$= \int_{\mathbf{P} \in \mathcal{P}^n} \Pr \left\{ \mathbf{z} \in (X \times Y)^{(n,m)} : \sup_{\mathbb{H}_l^n} d_\nu \left[\acute{\mathrm{er}}_{\mathbf{z}}(\mathbf{h}), \mathrm{er}_{\mathbf{P}}(\mathbf{h})\right] > \alpha \right\} \, dQ^n(\mathbf{P}). \quad (65)$$

Now $\mathbb{H}_l^n \subseteq K \oplus \cdots \oplus K$ where $K := \{h_l : h \in \mathcal{H} : \mathcal{H} \in \mathbb{H}\}$ and $\mathbb{H}_l^n$ is permissible by the assumed permissibility of $\mathbb{H}$ (Lemma 32, Appendix D). Hence $\mathbb{H}_l^n$ satisfies the conditions of Corollary 19 and so combining Lemma 24, Equation (65) and substituting $\alpha/2$ for $\alpha$ and $\delta/2$ for $\delta$ in Corollary 19 gives the following Lemma on the sample size required to ensure (62) holds.

**Lemma 25.** *If*

$$m \ge \max \left\{ \frac{32}{\alpha^2 \nu n} \log \frac{8 \mathcal{C}(\alpha \nu / 16, \mathbb{H}_l^n)}{\delta}, \frac{8}{\alpha^2 \nu} \right\}$$

*then*

$$\Pr \left\{ (\mathbf{z}, \mathbf{P}) \in (X \times Y)^{(n,m)} \times \mathcal{P}^n : \sup_{\mathbb{H}} d_\nu \left[\acute{\mathrm{er}}_{\mathbf{z}}(\mathcal{H}), \acute{\mathrm{er}}_{\mathbf{P}}(\mathcal{H})\right] > \frac{\alpha}{2} \right\} \le \frac{\delta}{2}.$$

### A.2.2 INEQUALITY (63)

Note that $\mathrm{er}_{\mathbf{P}}(\mathcal{H}) = \frac{1}{n} \sum_{i=1}^{n} \mathcal{H}^*(P_i)$ and $\mathrm{er}_Q(\mathcal{H}) = \mathbb{E}_{P \sim Q} \mathcal{H}^*(P)$, i.e the expectation of $\mathcal{H}^*(P)$ where $P$ is distributed according to $Q$. So to bound the left-hand-side of (63) we can apply Corollary 19 with $n = 1$, $m$ replaced by $n$, $\mathcal{H}$ replaced by $\mathbb{H}^*$, $\alpha$ and $\delta$ replaced by $\alpha/2$ and $\delta/2$ respectively, $P$ replaced by $Q$ and $Z$ replaced by $\mathcal{P}$. Note that $\mathbb{H}^*$ is permissible whenever $\mathbb{H}$ is (Lemma 32). Thus, if

$$n \ge \max \left\{ \frac{32}{\alpha^2 \nu} \log \frac{8 \mathcal{C}(\alpha \nu / 16, \mathbb{H}^*)}{\delta}, \frac{8}{\alpha^2 \nu} \right\} \quad (66)$$

then inequality (63) is satisfied.

Now, putting together Lemma 23, Lemma 25 and Equation 66, we have proved the following more general version of Theorem 2.

**Theorem 26.** *Let $\mathbb{H}$ be a permissible hypothesis space family and let $\mathbf{z}$ be an $(n, m)$-sample generated from the environment $(\mathcal{P}, Q)$. For all $0 < \alpha, \delta < 1$ and $\nu > 0$, if*

$$n \geq \max\left\{\frac{32}{\alpha^2\nu}\log\frac{8\mathcal{C}(\alpha\nu/16, \mathbb{H}^*)}{\delta}, \frac{8}{\alpha^2\nu}\right\}$$

$$and \qquad m \geq \max\left\{\frac{32}{\alpha^2\nu n}\log\frac{8\mathcal{C}(\alpha\nu/16, \mathbb{H}_l^n)}{\delta}, \frac{8}{\alpha^2\nu}\right\},$$

*then*

$$\Pr\left\{\mathbf{z} \in (X \times Y)^{(n,m)}\colon \sup_{\mathbb{H}} d_\nu\left[\hat{\mathrm{er}}_{\mathbf{z}}(\mathcal{H}), \mathrm{er}_Q(\mathcal{H})\right] > \alpha\right\} \leq \delta$$

To get Theorem 2, observe that $\mathrm{er}_Q(\mathcal{H}) > \hat{\mathrm{er}}_{\mathbf{z}}(\mathcal{H}) + \varepsilon \Rightarrow d_\nu\left[\hat{\mathrm{er}}_{\mathbf{z}}(\mathcal{H}), \mathrm{er}_Q(\mathcal{H})\right] > \varepsilon/(2 + \nu)$. Setting $\alpha = \varepsilon/(2 + \nu)$ and maximizing $\alpha^2\nu$ gives $\nu = 2$. Substituting $\alpha = \varepsilon/4$ and $\nu = 2$ into Theorem 26 gives Theorem 2.

### A.3 The Realizable Case

In Theorem 2 the sample complexity for both $m$ and $n$ scales as $1/\varepsilon^2$. This can be improved to $1/\varepsilon$ if instead of requiring $\mathrm{er}_Q(\mathcal{H}) \leq \hat{\mathrm{er}}_{\mathbf{z}}(\mathcal{H}) + \varepsilon$, we require only that $\mathrm{er}_Q(\mathcal{H}) \leq \kappa\hat{\mathrm{er}}_{\mathbf{z}}(\mathcal{H}) + \varepsilon$ for some $\kappa > 1$. To see this, observe that $\mathrm{er}_Q(\mathcal{H}) > \hat{\mathrm{er}}_{\mathbf{z}}(\mathcal{H})(1 + \alpha)/(1 - \alpha) + \alpha\nu/(1 - \alpha) \Rightarrow d_\nu\left[\hat{\mathrm{er}}_{\mathbf{z}}(\mathcal{H}), \mathrm{er}_Q(\mathcal{H})\right] > \alpha$, so setting $\alpha\nu/(1 - \alpha) = \varepsilon$ in Theorem 26 and treating $\alpha$ as a constant gives:

**Corollary 27.** *Under the same conditions as Theorem 26, for all $\varepsilon > 0$ and $0 < \alpha, \delta < 1$, if*

$$n \geq \max\left\{\frac{32}{\alpha(1-\alpha)\varepsilon}\log\frac{8\mathcal{C}((1-\alpha)\varepsilon/16, \mathbb{H}^*)}{\delta}, \frac{8}{\alpha(1-\alpha)\varepsilon}\right\}$$

$$and \qquad m \geq \max\left\{\frac{32}{\alpha(1-\alpha)\varepsilon n}\log\frac{8\mathcal{C}((1-\alpha)\varepsilon/16, \mathbb{H}_l^n)}{\delta}, \frac{8}{\alpha(1-\alpha)\varepsilon}\right\},$$

*then*

$$\Pr\left\{\mathbf{z} \in (X \times Y)^{(n,m)}\colon \sup_{\mathbb{H}} \mathrm{er}_Q(\mathcal{H}) \geq \frac{1+\alpha}{1-\alpha}\hat{\mathrm{er}}_{\mathbf{z}}(\mathcal{H}) + \varepsilon\right\} \leq \delta.$$

These bounds are particularly useful if we know that $\hat{\mathrm{er}}_{\mathbf{z}}(\mathcal{H}) = 0$, for then we can set $\alpha = 1/2$ (which maximizes $\alpha(1 - \alpha)$).

## Appendix B. Proof of Theorem 6

Recalling Definition 6, for $\mathbb{H}$ of the form given in (32), $\mathbb{H}_l^n$ can be written

$$\mathbb{H}_l^n = \{g_1 \circ f \oplus \cdots \oplus g_n \circ f\colon g_1, \ldots, g_n \in \mathcal{G}_l \text{ and } f \in \mathcal{F}\}.$$

To write $\mathbb{H}_l^n$ as a composition of two function classes note that if for each $f\colon X \to V$ we define $\bar{f}\colon (X \times Y)^n \to (V \times Y)^n$ by

$$\bar{f}(x_1, y_1, \ldots, x_n, y_n) := (f(x_1), y_1, \ldots, f(x_n), y_n)$$

then $g_1 \circ f \oplus \cdots \oplus g_n \circ f = g_1 \oplus \cdots \oplus g_n \circ \bar{f}$. Thus, setting $\mathcal{G}_l^n := \mathcal{G}_l \oplus \cdots \oplus \mathcal{G}_l$ and $\overline{\mathcal{F}} := \{\bar{f} : f \in \mathcal{F}\}$,

$$\mathbb{H}_l^n = \mathcal{G}_l^n \circ \overline{\mathcal{F}}. \tag{67}$$

The following two Lemmas will enable us to bound $\mathcal{C}\left(\varepsilon, \mathbb{H}_l^n\right)$.

**Lemma 28.** *Let* $\mathcal{H} \colon X \times Y \to [0,1]$ *be of the form* $\mathcal{H} = \mathcal{G}_l \circ \mathcal{F}$ *where* $X \times Y \xrightarrow{\mathcal{F}} V \times Y \xrightarrow{\mathcal{G}_l} [0,1]$. *For all* $\varepsilon_1, \varepsilon_2 > 0$,

$$\mathcal{C}(\varepsilon_1 + \varepsilon_2, \mathcal{H}) \leq \mathcal{C}_{\mathcal{G}_l}(\varepsilon_1, \mathcal{F}) \, \mathcal{C}(\varepsilon_2, \mathcal{G}_l).$$

*Proof.* Fix a measure $P$ on $X \times Y$ and let $F$ be a minimum size $\varepsilon_1$-cover for $(\mathcal{F}, d_{[P, \mathcal{G}_l]})$. By definition $|F| \leq \mathcal{C}_{\mathcal{G}_l}(\varepsilon_1, \mathcal{F})$. For each $f \in F$ let $P_f$ be the measure on $V \times Y$ defined by $P_f(S) = P(f^{-1}(S))$ for any set $S$ in the $\sigma$-algebra on $V \times Y$ ($f$ is measurable so $f^{-1}(S)$ is measurable). Let $G_f$ be a minimum size $\varepsilon_2$-cover for $(\mathcal{G}_l, d_{P_f})$. By definition again, $|G_f| \leq \mathcal{C}(\varepsilon_2, \mathcal{G}_l)$. Let $N := \{g \circ f : f \in F \text{ and } g \in G_f\}$. Note that $|N| \leq \mathcal{C}_{\mathcal{G}_l}(\varepsilon_1, \mathcal{F})\mathcal{C}(\varepsilon_2, \mathcal{G}_l)$ so the Lemma will be proved if $N$ can be shown to be an $\varepsilon_1 + \varepsilon_2$-cover for $(\mathcal{H}, d_P)$. So, given any $g \circ f \in \mathcal{H}$ choose $f' \in F$ such that $d_{[P, \mathcal{G}_l]}(f, f') \leq \varepsilon_1$ and $g' \in G_{f'}$ such that $d_{P_{f'}}(g, g') \leq \varepsilon_2$. Now,

$$\begin{aligned}
d_P(g \circ f, g' \circ f') &\leq d_P(g \circ f, g \circ f') + d_P(g \circ f', g' \circ f') \\
&\leq d_{[P, \mathcal{G}_l]}(f, f') + d_{P_{f'}}(g, g') \\
&\leq \varepsilon_1 + \varepsilon_2.
\end{aligned}$$

where the first line follows from the triangle inequality for $d_P$ and the second line follows from the facts: $d_P(g \circ f', g' \circ f') = d_{P_{f'}}(g, g')$ and $d_P(g \circ f, g \circ f') \leq d_{[P, \mathcal{G}_l]}(f, f')$. Thus $N$ is an $\varepsilon_1 + \varepsilon_2$-cover for $(\mathcal{H}, d_P)$ and so the result follows. $\qquad\square$

Recalling the definition of $\mathcal{H}_1 \oplus \cdots \oplus \mathcal{H}_n$ (Definition 6), we have the following Lemma.

**Lemma 29.**

$$\mathcal{C}(\varepsilon, \mathcal{H}_1 \oplus \cdots \oplus \mathcal{H}_n) \leq \prod_{i=1}^{n} \mathcal{C}(\varepsilon, \mathcal{H}_i)$$

*Proof.* Fix a product probability measure $\mathbf{P} = P_1 \times \cdots \times P_n$ on $(X \times Y)^n$. Let $N_1, \ldots, N_n$ be $\varepsilon$-covers of $(\mathcal{H}_1, d_{P_1}) \ldots, (\mathcal{H}_n, d_{P_n})$. and let $N = N_1 \oplus \cdots \oplus N_n$. Given $h = h_1 \oplus \cdots \oplus h_n \in \mathcal{H}_1 \oplus \cdots \oplus \mathcal{H}_n$, choose $g_1 \oplus \cdots \oplus g_n \in N$ such that $d_{P_i}(h_i, g_i) \leq \varepsilon$ for each $i = 1, \ldots, n$. Now,

$$\begin{aligned}
d_{\mathbf{P}}(h_1 \oplus \cdots \oplus h_n, g_1 \oplus \cdots \oplus g_n) &= \frac{1}{n} \int_{Z^n} \left| \sum_{i=1}^{n} h_i(z_i) - \sum_{i=1}^{n} g_i(z_i) \right| d\mathbf{P}(z_1, \ldots, z_n) \\
&\leq \frac{1}{n} \sum_{i=1}^{n} d_{P_i}(h_i, g_i) \\
&\leq \varepsilon.
\end{aligned}$$

Thus $N$ is an $\varepsilon$-cover for $\mathcal{H}_1 \oplus \cdots \oplus \mathcal{H}_n$ and as $|N| = \prod_{i=1}^{n} |N_i|$ the result follows. $\qquad\square$

## B.1  Bounding $\mathcal{C}\left(\varepsilon, \mathbb{H}_l^n\right)$

From Lemma 28,

$$\mathcal{C}\left(\varepsilon_1 + \varepsilon_2, \mathcal{G}_l{}^n \circ \overline{\mathcal{F}}\right) \leq \mathcal{C}\left(\varepsilon_1, \mathcal{G}_l{}^n\right) \mathcal{C}_{\mathcal{G}_l{}^n}\left(\varepsilon_2, \overline{\mathcal{F}}\right) \tag{68}$$

and from Lemma 29,

$$\mathcal{C}\left(\varepsilon_1, \mathcal{G}_l{}^n\right) \leq \mathcal{C}\left(\varepsilon_1, \mathcal{G}_l\right)^n. \tag{69}$$

Using similar techniques to those used to prove Lemmas 28 and 29, $\mathcal{C}_{\mathcal{G}_l{}^n}(\varepsilon, \overline{\mathcal{F}})$ can be shown to satisfy

$$\mathcal{C}_{\mathcal{G}_l{}^n}\left(\varepsilon_2, \overline{\mathcal{F}}\right) \leq \mathcal{C}_{\mathcal{G}_l}\left(\varepsilon_2, \mathcal{F}\right). \tag{70}$$

Equations (67), (68), (69) and (70) together imply inequality (34).

## B.2  Bounding $\mathcal{C}\left(\varepsilon, \mathbb{H}^*\right)$

We wish to prove that $\mathcal{C}\left(\varepsilon, \mathbb{H}^*\right) \leq \mathcal{C}_{\mathcal{G}_l}\left(\varepsilon, \mathcal{F}\right)$ when $\mathbb{H}$ is a hypothesis space family of the form $\mathbb{H} = \{\mathcal{G}_l \circ f : f \in \mathcal{F}\}$. Note that each $\mathcal{H}^* \in \mathbb{H}^*$ corresponds to some $\mathcal{G}_l \circ f$, and that

$$\mathcal{H}^*(P) = \inf_{g \in \mathcal{G}_l} \operatorname{er}_P(g \circ f).$$

Any probability measure $Q$ on $\mathcal{P}$ induces a probability measure $Q_{X \times Y}$ on $X \times Y$, defined by

$$Q_{X \times Y}(S) = \int_{\mathcal{P}} P(S)\, dQ(P)$$

for any $S$ in the $\sigma$-algebra on $X \times Y$. Note also that if $h, h'$ are bounded, positive functions on an arbitrary set $A$, then

$$\left| \inf_{a \in A} h(a) - \inf_{a \in A} h'(a) \right| \leq \sup_{a \in A} \left| h(a) - h'(a) \right|. \tag{71}$$

Let $Q$ be any probability measure on the space $\mathcal{P}$ of probability measures on $X \times Y$. Let $\mathcal{H}_1^*, \mathcal{H}_2^*$ be two elements of $\mathbb{H}^*$ with corresponding hypothesis spaces $\mathcal{G}_l \circ f_1, \mathcal{G}_l \circ f_2$. Then,

$$
\begin{aligned}
d_Q(\mathcal{H}_1^*, \mathcal{H}_2^*) &= \int_{\mathcal{P}} \left| \inf_{g \in \mathcal{G}_l} \operatorname{er}_P(g \circ f_1) - \inf_{g \in \mathcal{G}_l} \operatorname{er}_P(g \circ f_2) \right| dQ(P) \\
&\leq \int_{\mathcal{P}} \sup_{g \in \mathcal{G}_l} |\operatorname{er}_P(g \circ f_1) - \operatorname{er}_P(g \circ f_2)|\ dQ(P) \qquad \text{(by (71) above)} \\
&\leq \int_{\mathcal{P}} \int_{X \times Y} \sup_{g \in \mathcal{G}_l} |g \circ f_1(x, y) - g \circ f_2(x, y)|\ dP(x, y)\, dQ(P) \\
&= d_{\lfloor Q_{X \times Y}, \mathcal{G}_l \rfloor}(f_1, f_2).
\end{aligned}
$$

The measurability of $\sup_{\mathcal{G}_l} g \circ f$ is guaranteed by the permissibility of $\mathbb{H}$ (Lemma 32 part 4, Appendix D). From $d_Q(\mathcal{H}_1^*, \mathcal{H}_2^*) \leq d_{\lfloor Q_{X \times Y}, \mathcal{G}_l \rfloor}(f_1, f_2)$ we have,

$$\mathcal{N}\left(\varepsilon, \mathbb{H}^*, d_Q\right) \leq \mathcal{N}\left(\varepsilon, \mathcal{F}, d_{\lfloor Q_{X \times Y}, \mathcal{G}_l \rfloor}\right), \tag{72}$$

which gives inequality (35).

### B.3 Proof of Theorem 7

In order to prove the bounds in Theorem 7 we have to apply Theorem 6 to the neural network hypothesis space family of equation (39). In this case the structure is

$$\mathbb{R}^d \xrightarrow{\mathcal{F}} \mathbb{R}^k \xrightarrow{\mathcal{G}} [0,1]$$

where $\mathcal{G} = \left\{ (x_1, \ldots, x_k) \mapsto \sigma \left( \sum_{i=1}^k \alpha_i x_i + \alpha_0 \right) : (\alpha_0, \alpha_1, \ldots, \alpha_k) \in U \right\}$ for some bounded subset $U$ of $\mathbb{R}^{k+1}$ and some Lipschitz squashing function $\sigma$. The feature class $\mathcal{F} \colon \mathbb{R}^d \to \mathbb{R}^k$ is the set of all one hidden layer neural networks with $d$ inputs, $l$ hidden nodes, $k$ outputs, $\sigma$ as the squashing function and weights $w \in T$ where $T$ is a bounded subset of $\mathbb{R}^W$. The Lipschitz restriction on $\sigma$ and the bounded restrictions on the weights ensure that $\mathcal{F}$ and $\mathcal{G}$ are Lipschitz classes. Hence there exists $b < \infty$ such that for all $f \in \mathcal{F}$ and $x, x' \in \mathbb{R}^d$, $\|f(x) - f(x')\| < b\|x - x'\|$ and for all $g \in \mathcal{G}$ and $x, x' \in \mathbb{R}^k$, $|g(x) - g(x')| < b\|x - x'\|$ where $\|\cdot\|$ is the $L_1$ norm in each case. The loss function is squared loss.

Now, $g_l(x, y) = l(g(x), y) = (g(x) - y)^2$, hence for all $g, g' \in \mathcal{G}$ and all probability measures $P$ on $\mathbb{R}^k \times [0, 1]$ (recall that we assumed the output space $Y$ was $[0,1]$),

$$
\begin{aligned}
d_P(g_l, g_l') &= \int_{\mathbb{R}^k \times [0,1]} \left| (g(v) - y)^2 - (g'(v) - y)^2 \right| \, dP(v, y) \\
&\leq 2 \int_{\mathbb{R}^k} \left| g(v) - g'(v) \right| \, dP_{\mathbb{R}^k}(v),
\end{aligned}
\tag{73}
$$

where $P_{\mathbb{R}^k}$ is the marginal distribution on $\mathbb{R}^k$ derived from $P$. Similarly, for all $f, f' \in \mathcal{F}$ and probability measures $P$ on $\mathbb{R}^d \times [0, 1]$,

$$
d_{[P, \mathcal{G}_l]}(f, f') \leq 2b \int_{\mathbb{R}^d} \|f(x) - f'(x)\| \, dP_{\mathbb{R}^d}(x).
\tag{74}
$$

Define

$$
\mathcal{C}\left( \varepsilon, \mathcal{G}, L^1 \right) := \sup_P \mathcal{N}\left( \varepsilon, \mathcal{G}, L^1(P) \right),
$$

where the supremum is over all probability measures on (the Borel subsets of) $\mathbb{R}^k$, and $\mathcal{N}\left( \varepsilon, \mathcal{G}, L^1(P) \right)$ is the size of the smallest $\varepsilon$-cover of $\mathcal{G}$ under the $L^1(P)$ metric. Similarly set,

$$
\mathcal{C}\left( \varepsilon, \mathcal{F}, L^1 \right) := \sup_P \mathcal{N}\left( \varepsilon, \mathcal{F}, L^1(P) \right),
$$

where now the supremum is over all probability measures on $\mathbb{R}^d$. Equations (73) and (74) imply

$$
\mathcal{C}(\varepsilon, \mathcal{G}_l) \leq \mathcal{C}\left( \frac{\varepsilon}{2}, \mathcal{G}, L^1 \right)
\tag{75}
$$

$$
\mathcal{C}_{\mathcal{G}_l}\left( \varepsilon, \mathcal{F} \right) \leq \mathcal{C}\left( \frac{\varepsilon}{2b}, \mathcal{F}, L^1 \right)
\tag{76}
$$

Applying Theorem 11 from Haussler (1992), we find

$$
\mathcal{C}\left( \frac{\varepsilon}{2}, \mathcal{G}_l, L^1 \right) \leq \left[ \frac{2eb}{\varepsilon} \right]^{2k+2}
$$

$$
\mathcal{C}\left( \frac{\varepsilon}{2b}, \mathcal{F}, L^1 \right) \leq \left[ \frac{2eb^2}{\varepsilon} \right]^{2W}.
$$

Substituting these two expressions into (75) and (76) and applying Theorem 6 yields Theorem 7. □

## Appendix C. Proof of Theorem 14

This proof follows a similar argument to the one presented in Anthony and Bartlett (1999) for ordinary Boolean function learning.

First we need a technical Lemma.

**Lemma 30.** *Let $\alpha$ be a random variable uniformly distributed on $\{1/2 + \beta/2, 1/2 - \beta/2\}$, with $0 < \beta < 1$. Let $\xi_1, \ldots, \xi_m$ be i.i.d. $\{1, -1\}$-valued random variables with $\Pr(\xi_i = 1) = \alpha$ for all $i$. For any function $f$ mapping $\{1, -1\}^n \to \{1/2 + \beta/2, 1/2 - \beta/2\}$,*

$$\Pr\{\xi_1, \ldots, \xi_m \colon f(\xi_1, \ldots, \xi_m) \neq \alpha\} > \frac{1}{4}\left[1 - \sqrt{1 - e^{-\frac{m\beta^2}{1-\beta^2}}}\right].$$

*Proof.* Let $N(\xi)$ denote the number of occurences of $+1$ in the random sequence $\xi = (\xi_1, \ldots, \xi_m)$. The function $f$ can be viewed as a decision rule, i.e. based on the observations $\xi$, $f$ tries to guess whether the probability of $+1$ is $1/2 + \beta/2$ or $1/2 - \beta/2$. The optimal decision rule is the Bayes estimator: $f(\xi_1, \ldots, \xi_m) = 1/2 + \beta/2$ if $N(\xi) \geq m/2$, and $f(\xi_1, \ldots, \xi_m) = 1/2 - \beta/2$ otherwise. Hence,

$$\begin{aligned}
\Pr(f(\xi) \neq \alpha) &\geq \frac{1}{2}\Pr\left(N(\xi) \geq \frac{m}{2} \,\Big|\, \alpha = \frac{1}{2} - \frac{\beta}{2}\right) \\
&\quad + \frac{1}{2}\Pr\left(N(\xi) < \frac{m}{2} \,\Big|\, \alpha = \frac{1}{2} + \frac{\beta}{2}\right) \\
&> \frac{1}{2}\Pr\left(N(\xi) \geq \frac{m}{2} \,\Big|\, \alpha = \frac{1}{2} - \frac{\beta}{2}\right)
\end{aligned}$$

which is half the probability that a binomial $(m, 1/2 - \beta/2)$ random variable is at least $m/2$. By Slud's inequality (Slud, 1977),

$$\Pr(f(\xi) \neq \alpha) > \frac{1}{2}\Pr\left(Z \geq \sqrt{\frac{m\beta^2}{1 - \beta^2}}\right)$$

where $Z$ is normal $(0, 1)$. Tate's inequality (Tate, 1953) states that for all $x \geq 0$,

$$\Pr(Z \geq x) \geq \frac{1}{2}\left[1 - \sqrt{1 - e^{-x^2}}\right].$$

Combining the last two inequalities completes the proof. $\qquad\square$

Let $\mathbf{x} \in X^{(n,m)}$ be shattered by $\mathbb{H}$, with $m = d_{\mathbb{H}}(n)$. For each row $i$ in $\mathbf{x}$ let $\mathcal{P}_i$ be the set of all $2^d$ distributions $P$ on $X \times \{\pm 1\}$ such that $P(x, 1) = P(x, 0) = 0$ if $x$ is not contained in the $i$th row of $\mathbf{x}$, and for each $j = 1, \ldots, d_{\mathbb{H}}(n)$, $P(x_{ij}, 1) = (1 \pm \beta)/(2d_{\mathbb{H}}(n))$ and $P(x_{ij}, -1) = (1 \mp \beta)/(2d_{\mathbb{H}}(n))$. Let $\mathcal{P} := \mathcal{P}_1 \times \cdots \times \mathcal{P}_n$.

Note that for $\mathbf{P} = (P_1, \ldots, P_n) \in \mathcal{P}$, the optimal error $\mathrm{opt}_{\mathbf{P}}(\mathbb{H}^n)$ is achieved by any sequence $\mathbf{h}^* = (h_1^*, \ldots, h_n^*)$ such that $h_i^*(x_{ij}) = 1$ if and only if $P_i(x_{ij}, 1) = (1 + \beta)/(2d_{\mathbb{H}}(n))$, and $\mathbb{H}^n$ always contains such a sequence because $\mathbb{H}$ shatters $\mathbf{x}$. The optimal error is then

$$\mathrm{opt}_{\mathbf{P}}(\mathbb{H}^n) = \mathrm{er}_{\mathbf{P}}(\mathbf{h}^*) = \frac{1}{n}\sum_{i=1}^{n} P_i\{h_i^*(x) \neq y\} = \frac{1}{n}\sum_{i=1}^{n}\sum_{j=1}^{d_{\mathbb{H}}(n)} \frac{1 - \beta}{2d_{\mathbb{H}}(n)} = \frac{1 - \beta}{2},$$

and for any $\mathbf{h} = (h_1, \ldots, h_n) \in \mathbb{H}^n$,

$$\text{er}_{\mathbf{P}}(\mathbf{h}) = \text{opt}_{\mathbf{P}}(\mathbb{H}^n) + \frac{\beta}{n d_{\mathbb{H}}(n)} |\{(i, j) \colon h_i(x_{ij}) \neq h_i^*(x_{ij})\}|. \tag{77}$$

For any $(n, m)$-sample $\mathbf{z}$, let each element $m_{ij}$ in the array

$$\mathbf{m}(\mathbf{z}) := \begin{matrix} m_{11} & \cdots & m_{1d_{\mathbb{H}}(n)} \\ \vdots & \ddots & \vdots \\ m_{n1} & \cdots & m_{nd_{\mathbb{H}}(n)} \end{matrix}$$

equal the number of occurrences of $x_{ij}$ in $\mathbf{z}$.

Now, if we select $\mathbf{P} = (P_1, \ldots, P_n)$ uniformly at random from $\mathcal{P}$, and generate an $(n, m)$-sample $\mathbf{z}$ using $\mathbf{P}$, then for $\mathbf{h} = \mathcal{A}_n(\mathbf{z})$ (the output of the learning algorithm) we have:

$$\mathbb{E}\left(|\{(i, j) \colon h_i(x_{ij}) \neq h_i^*(x_{ij})\}|\right) = \sum_{\mathbf{m}} P(\mathbf{m}) \mathbb{E}\left(|\{(i, j) \colon h_i(x_{ij}) \neq h_i^*(x_{ij})\}| \mid \mathbf{m}\right)$$

$$= \sum_{\mathbf{m}} P(\mathbf{m}) \sum_{i=1}^{n} \sum_{j=1}^{d_{\mathbb{H}}(n)} P\left(h(x_{ij}) \neq h^*(x_{ij}) \mid m_{ij}\right)$$

where $P(\mathbf{m})$ is the probability of generating a configuration $\mathbf{m}$ of the $x_{ij}$ under the $(n, m)$-sampling process and the sum is over all possible configurations. From Lemma 30,

$$P\left(h(x_{ij}) \neq h^*(x_{ij}) \mid m_{ij}\right) > \frac{1}{4}\left[1 - \sqrt{1 - e^{-\frac{m_{ij}\beta^2}{1-\beta^2}}}\right],$$

hence

$$\mathbb{E}\left[\frac{1}{n d_{\mathbb{H}}(n)} |\{(i, j) \colon h_i(x_{ij}) \neq h_i^*(x_{ij})\}|\right] > \frac{1}{n d_{\mathbb{H}}(n)} \sum_{\mathbf{m}} P(\mathbf{m}) \sum_{i=1}^{n} \sum_{j=1}^{d_{\mathbb{H}}(n)} \frac{1}{4}\left[1 - \sqrt{1 - e^{-\frac{m_{ij}\beta^2}{1-\beta^2}}}\right]$$

$$\geq \frac{1}{4}\left[1 - \sqrt{1 - e^{-\frac{m\beta^2}{d_{\mathbb{H}}(n)(1-\beta^2)}}}\right] \tag{78}$$

by Jensen's inequality. Since for any $[0, 1]$-valued random variable $Z$, $\Pr(Z > x) \geq \mathbb{E}Z - x$, (78) implies:

$$\Pr\left(\frac{1}{n d_{\mathbb{H}}(n)} |\{(i, j) \colon h_i(x_{ij}) \neq h_i^*(x_{ij})\}| > \gamma \alpha\right) > (1 - \gamma)\alpha$$

where

$$\alpha := \frac{1}{4}\left[1 - \sqrt{1 - e^{-\frac{m\beta^2}{d_{\mathbb{H}}(n)(1-\beta^2)}}}\right] \tag{79}$$

and $\gamma \in [0, 1]$. Plugging this into (77) shows that

$$\Pr\left\{(\mathbf{P}, \mathbf{z}) \colon \text{er}_{\mathbf{P}}(\mathcal{A}_n(\mathbf{z})) > \text{opt}_{\mathbf{P}}(\mathbb{H}^n) + \gamma \alpha \beta\right\} > (1 - \gamma)\alpha.$$

Since the inequality holds over the random choice of $\mathbf{P}$, it must also hold for some specific choice of $\mathbf{P}$. Hence for any learning algorithm $\mathcal{A}_n$ there is some sequence of distributions $\mathbf{P}$ such that

$$\Pr\left\{\mathbf{z}\colon \operatorname{er}_{\mathbf{P}}(\mathcal{A}_n(\mathbf{z})) > \operatorname{opt}_{\mathbf{P}}(\mathbb{H}^n) + \gamma\alpha\beta\right\} > (1-\gamma)\alpha.$$

Setting

$$(1-\gamma)\alpha \geq \delta, \quad \text{and} \quad \gamma\alpha\beta \geq \varepsilon, \tag{80}$$

ensures

$$\Pr\left\{\mathbf{z}\colon \operatorname{er}_{\mathbf{P}}(\mathcal{A}_n(\mathbf{z})) > \operatorname{opt}_{\mathbf{P}}(\mathbb{H}^n) + \varepsilon\right\} > \delta. \tag{81}$$

Assuming equality in (80), we get

$$\alpha = \frac{\delta}{1-\gamma}, \qquad \beta = \frac{\varepsilon}{\delta}\frac{1-\gamma}{\gamma}.$$

Solving (79) for $m$, and substituting the above expressions for $\alpha$ and $\beta$ shows that (81) is satisfied provided

$$m \leq d_{\mathbb{H}}(n)\left[\left(\frac{\delta}{\varepsilon}\right)^2\left(\frac{\gamma}{1-\gamma}\right)^2 - 1\right]\log\frac{(1-\gamma)^2}{8\delta\left(1-\gamma-2\delta\right)} \tag{82}$$

Setting $\gamma = 1 - a\delta$ for some $a > 4$ ($a > 4$ since $\alpha < 1/4$ and $\alpha = \delta/(1-\gamma)$), and assuming $\varepsilon, \delta \leq 1/(ka)$ for some $k > 2$, (82) becomes

$$m \leq \frac{d_{\mathbb{H}}(n)}{a^2}\left[1 - \frac{2}{k}\right]\log\frac{a^2}{8(a-2)}. \tag{83}$$

Subject to the constraint $a > 4$, the right hand side of (83) is approximately maximized at $a = 8.7966$, at which point the value exceeds $d_{\mathbb{H}}(n)(1 - 2/k)/(220\varepsilon^2)$. Thus, for all $k \geq 1$, if $\varepsilon, \delta \leq 1/9k$ and

$$m \leq \frac{d_{\mathbb{H}}(n)\left(1 - \frac{2}{k}\right)}{220\varepsilon^2}, \tag{84}$$

then

$$\Pr\left\{\mathbf{z}\colon \operatorname{er}_{\mathbf{P}}(\mathcal{A}_n(\mathbf{z})) > \operatorname{opt}_{\mathbf{P}}(\mathbb{H}^n) + \varepsilon\right\} > \delta.$$

To obtain the $\delta$-dependence in Theorem 14 observe that by assumption $\mathbb{H}^1$ contains at least two functions $h_1, h_2$, hence there exists an $x \in X$ such that $h_1(x) \neq h_2(x)$. Let $P^{\pm}$ be two distributions concentrated on $(x, 1)$ and $(x, -1)$ such that $P^{\pm}(x, h_1(x)) = (1 \pm \varepsilon)/2$ and $P^{\pm}(x, h_2(x)) = (1 \mp \varepsilon)/2$. Let $\mathbf{P}^+ := P^+ \times \cdots \times P^+$ and $\mathbf{P}^- := P^- \times \cdots \times P^-$ be the product distributions on $(X \times \{\pm 1\})^n$ generated by $P^{\pm}$, and $\mathbf{h}_1 := (h_1, \ldots, h_1), \mathbf{h}_2 := (h_2, \ldots, h_2)$. Note that $\mathbf{h}_1$ and $\mathbf{h}_2$ are both in $\mathbb{H}^n$. If $\mathbf{P}$ is one of $\mathbf{P}^{\pm}$ and the learning algorithm $\mathcal{A}_n$ chooses the wrong hypothesis $\mathbf{h}$, then

$$\operatorname{er}_{\mathbf{P}}(\mathbf{h}) - \operatorname{opt}_{\mathbf{P}}(\mathbb{H}^n) = \varepsilon.$$

Now, if we choose $\mathbf{P}$ uniformly at random from $\{\mathbf{P}^+, \mathbf{P}^-\}$ and generate an $(n, m)$-sample $\mathbf{z}$ according to $\mathbf{P}$, Lemma 30 shows that

$$\Pr\{(\mathbf{P}, \mathbf{z})\colon\ \mathrm{er}_{\mathbf{P}}(\mathcal{A}_n(\mathbf{z})) \geq \mathrm{opt}_{\mathbf{P}}(\mathbb{H}^n) + \varepsilon\} > \frac{1}{4}\left[1 - \sqrt{1 - e^{\frac{n m \varepsilon^2}{1 - \varepsilon^2}}}\right],$$

which is at least $\delta$ if

$$m < \frac{1 - \varepsilon^2}{\varepsilon^2} \frac{1}{n} \log \frac{1}{8\delta(1 - 2\delta)} \tag{85}$$

provided $0 < \delta < 1/4$. Combining the two constraints on $m$: (84) (with $k = 7$) and (85), and using $\max\{x_1, x_2\} \geq \frac{1}{2}(x_1 + x_2)$ finishes the proof. □

## Appendix D. Measurability

In order for Theorems 2 and 18 to hold in full generality we had to impose a constraint called "permissibility" on the hypothesis space family $\mathbb{H}$. Permissibility was introduced by Pollard (1984) for ordinary hypothesis classes $\mathcal{H}$. His definition is very similar to Dudley's "image admissible Suslin" (Dudley, 1984). We will be extending this definition to cover hypothesis space families.

Throughout this section we assume all functions $h$ map from (the complete separable metric space) $Z$ into $[0, 1]$. Let $\mathcal{B}(T)$ denote the Borel $\sigma$-algebra of any topological space $T$. As in Section 2.2, we view $\mathcal{P}$, the set of all probability measures on $Z$, as a topological space by equipping it with the topology of weak convergence. $\mathcal{B}(\mathcal{P})$ is then the $\sigma$-algebra generated by this topology. The following two definitions are taken (with minor modifications) from Pollard (1984).

**Definition 8.** *A set $\mathcal{H}$ of $[0, 1]$-valued functions on $Z$ is* indexed *by the set $T$ if there exists a function $f\colon Z \times T \to [0, 1]$ such that*

$$\mathcal{H} = \{f(\,\cdot\,, t)\colon t \in T\}.$$

**Definition 9.** *The set $\mathcal{H}$ is* permissible *if it can be indexed by a set $T$ such that*

1. *$T$ is an analytic subset of a Polish[7] space $\overline{T}$, and*

2. *the function $f\colon Z \times T \to [0, 1]$ indexing $\mathcal{H}$ by $T$ is measurable with respect to the product $\sigma$-algebra $\mathcal{B}(Z) \otimes \mathcal{B}(T)$.*

An analytic subset $T$ of a Polish space $\overline{T}$ is simply the continuous image of a Borel subset $X$ of another Polish space $\overline{X}$. The analytic subsets of a Polish space include the Borel sets. They are important because projections of analytic sets are analytic, and can be measured in a complete measure space whereas projections of Borel sets are not necessarily Borel, and hence cannot be measured with a Borel measure. For more details see Dudley (1989), section 13.2.

**Lemma 31.** *$\mathcal{H}_1 \oplus \cdots \oplus \mathcal{H}_n\colon (X \times Y)^n \to [0, 1]$ is permissible if $\mathcal{H}_1, \ldots, \mathcal{H}_n$ are all permissible.*

*Proof.* Omitted. ■

We now define permissibility of hypothesis space families.

---

7. A topological space is called *Polish* if it is metrizable such that it is a complete separable metric space.

**Definition 10.** *A hypothesis space family* $\mathbb{H} = \{\mathcal{H}\}$ *is* permissible *if there exist sets $S$ and $T$ that are analytic subsets of Polish spaces $\overline{S}$ and $\overline{T}$ respectively, and a function $f\colon Z \times T \times S \to [0,1]$, measurable with respect to $\mathcal{S} \otimes \mathcal{B}(T) \otimes \mathcal{B}(S)$, such that*

$$\mathbb{H} = \big\{ \{f(\,\cdot\,, t, s)\colon t \in T\}\colon s \in S \big\}.$$

Let $(X, \Sigma, \mu)$ be a measure space and $T$ be an analytic subset of a Polish space. Let $\mathcal{A}(X)$ denote the analytic subsets of $X$. The following three facts about analytic sets are taken from Pollard (1984), appendix C.

(a) If $(X, \Sigma, \mu)$ is complete then $\mathcal{A}(X) \subseteq \Sigma$.

(b) $\mathcal{A}(X \times T)$ contains the product $\sigma$-algebra $\Sigma \otimes \mathcal{B}(T)$.

(c) For any set $Y$ in $\mathcal{A}(X \times T)$, the projection $\pi_X Y$ of $Y$ onto $X$ is in $\mathcal{A}(X)$.

Recall Definition 2 for the definition of $\mathbb{H}^*$. In the following Lemma we assume that $(Z, \mathcal{B}(Z))$ has been completed with respect to any probability measure $P$, and also that $(\mathcal{P}, \mathcal{B}(\mathcal{P}))$ is complete with respect to the environmental measure $Q$.

**Lemma 32.** *For any permissible hypothesis space family $\mathbb{H}$,*

1. *$\mathbb{H}_l^n$ is permissible.*

2. *$\{h \in \mathcal{H}\colon \mathcal{H} \in \mathbb{H}\}$ is permissible.*

3. *$\mathcal{H}$ is permissible for all $\mathcal{H} \in \mathbb{H}$.*

4. *$\sup_{\mathcal{H}}$ and $\inf_{\mathcal{H}}$ are measurable for all $\mathcal{H} \in \mathbb{H}$.*

5. *$\mathcal{H}^*$ is measurable for all $\mathcal{H} \in \mathbb{H}$.*

6. *$\mathbb{H}^*$ is permissible.*

*Proof.* As we have absorbed the loss function into the hypotheses $h$, $\mathbb{H}_l^n$ is simply the set of all $n$-fold products $\mathcal{H} \oplus \cdots \oplus \mathcal{H}$ such that $\mathcal{H} \in \mathbb{H}$. Thus (1) follows from Lemma 31. (2) and (3) are immediate from the definitions. As $\mathcal{H}$ is permissible for all $\mathcal{H} \in \mathbb{H}$, (4) can be proved by an identical argument to that used in the "Measurable Suprema" section of Pollard (1984), appendix C.

For (5), note that for any Borel-measurable $h\colon Z \to [0,1]$, the function $\overline{h}\colon \mathcal{P} \to [0,1]$ defined by $\overline{h}(P) := \int_Z h(z)\, dP(z)$ is Borel measurable Kechris (1995, chapter 17). Now, permissibility of $\mathcal{H}$ automatically implies permissibility of $\overline{\mathcal{H}} := \{\overline{h}\colon h \in \mathcal{H}\}$, and $\mathcal{H}^* = \inf_{\overline{\mathcal{H}}}$ so $\mathcal{H}^*$ is measurable by (4).

Now let $\mathbb{H}$ be indexed by $f\colon Z \times T \times S \to [0,1]$ in the appropriate way. To prove (6), define $g\colon \mathcal{P} \times T \times S \to [0,1]$ by $g(P, t, s) := \int_Z f(z, t, s)\, dP(z)$. By Fubini's theorem $g$ is a $\mathcal{B}(\mathcal{P}) \otimes \mathcal{B}(T) \otimes \mathcal{B}(S)$-measurable function. Let $G\colon \mathcal{P} \times S \to [0,1]$ be defined by $G(P, s) := \inf_{t \in T} g(P, t, s)$. $G$ indexes $\mathbb{H}^*$ in the appropriate way for $\mathbb{H}^*$ to be permissible, provided it can be shown that $G$ is $\mathcal{B}(\mathcal{P}) \otimes \mathcal{B}(S)$-measurable. This is where analyticity becomes important. Let $g_\alpha := \{(P, t, s)\colon g(P, t, s) > \alpha\}$. By property (b) of analytic sets, $\mathcal{A}(\mathcal{P} \times T \times S)$ contains $g_\alpha$. The set $G_\alpha := \{(P, s)\colon G(P, s) > \alpha\}$ is the projection of $g_\alpha$ onto $\mathcal{P} \times S$, which by property (c) is also analytic. As $(\mathcal{P}, \mathcal{B}(\mathcal{P}), Q)$ is assumed complete, $G_\alpha$ is measurable, by property (a). Thus $G$ is a measurable function and the permissibility of $\mathbb{H}^*$ follows. $\qquad\square$